\documentclass{article}

\PassOptionsToPackage{numbers, sort}{natbib}
 \usepackage[preprint]{neurips_2026}

\usepackage[utf8]{inputenc} % allow utf-8 input
\usepackage[T1]{fontenc}    % use 8-bit T1 fonts
\usepackage{hyperref}       % hyperlinks
\usepackage{url}            % simple URL typesetting
\usepackage{booktabs}       % professional-quality tables
\usepackage{amsfonts}       % blackboard math symbols
\usepackage{nicefrac}       % compact symbols for 1/2, etc.
\usepackage{microtype}      % microtypography
\usepackage{xcolor}         % colors
\usepackage{graphicx}
\usepackage{amsmath}
\usepackage{cleveref}
\usepackage{mathtools}
\usepackage{verbatim}
\usepackage{multirow}
\usepackage{pifont}
\usepackage{caption}
\usepackage{wrapfig}

\usepackage{booktabs}
\usepackage[table]{xcolor}
\usepackage{multirow}
\usepackage{amssymb}
\usepackage{makecell}
\usepackage{xfrac} % optional, only if you want \sfrac
\usepackage{algorithm}
\usepackage{algorithmic}

\definecolor{targetblue}{RGB}{232,242,255}
\definecolor{oursblue}{RGB}{214,232,255}

\newcommand{\cmark}{\checkmark}

\newcommand{\blfootnote}[1]{%
  \begingroup
  \renewcommand{\thefootnote}{}%
  \footnotetext{#1}%
  \endgroup
}
\hypersetup{
    colorlinks=true,
    urlcolor=blue,
    linkcolor=black,
    citecolor=black
}
\title{Constrained Goal-directed Planar Graph Generation with Grammar-based Reinforcement Learning}

\author{%
  Nicolas Hochuli\thanks{Equal contribution. Correspondence to \texttt{nhochuli@ethz.ch}}\\
  Engineering Design and\\
   Computing Laboratory\\
  ETH Zürich\\
  \And
  Lorenzo Miele$^\ast$\\
  Engineering Design and\\
   Computing Laboratory\\
  ETH Zürich\\
  \And
    Kristina Shea \\
  Engineering Design and\\
   Computing Laboratory\\
  ETH Zürich\\
  \And
    Tino Stankovic\\
  Engineering Design and\\
   Computing Laboratory\\
  ETH Zürich\\
}

\begin{document}

\maketitle
\blfootnote{Code is available at
\url{https://github.com/ETHZ-EDAC/constrained-graph-rl}.}
%Planar graphs are central to many applications, yet existing approaches offer limited support for goal-directed graph generation while adhering to valid planar embeddings. In response, we propose a dataset-free method for generating planar graph embeddings that integrates parametric graph grammars with safe reinforcement learning, enabling simultaneous optimization of structural, geometric, and spectral objectives. We formulate the graph generation as a constrained Markov decision process, and introduce a differentiable action projection to facilitate constraint satisfaction throughout the learning process. Unlike classical or distribution-matching generators, the planar embedding is constructed during generation while aligning with user-specified multi-objective goals, without relying on a dataset and enforcing constraints during generation. We further introduce dedicated benchmarks and baseline models for these settings and demonstrate improved scalability and performance on these tasks.
\begin{abstract}
Planar graphs are central to applications across science and engineering, yet existing generators provide limited support for goal-directed generation under hard structural and geometric feasibility constraints. We propose a dataset-free method for generating planar graph embeddings by combining parametric graph grammars with safe reinforcement learning to optimize generic task-specific objectives while satisfying constraints \textit{during} construction. We formulate the generation process as a constrained Markov decision process, where the graph grammar defines the state and action spaces. We further introduce an action projection that maps sampled actions toward state-dependent safe sets, improving constraint satisfaction during training. In contrast to classical graph generators and deep generative models, which typically offer limited goal-directed control or rely on weak constraint satisfaction, our method constructs feasible planar graph embeddings directly during generation. We also introduce a benchmark suite for constrained and goal-directed planar graph generation, together with classical and deep generative baselines. Across all benchmark tasks, our method consistently outperforms baselines while satisfying the formulated constraints.
\end{abstract}

\section{Introduction}
% Goal-directed graph generation is a central research topic in computer science, with applications across domains ranging from complex networks, bioinformatics, and urban planning~\citep{barthelemy_spatial_2011, you_graph_2018, boeing_osmnx_2017} to engineering design, including architecture, structural optimization, and origami design~\citep{kakooee_reimagining_2024, ororbia_design_2021, walker_algorithmic_2022}
Goal-directed planar graph generation is a central research topic in computer science, with applications spanning urban planning and road-network design ~\citep{barthelemy_spatial_2011, you_graph_2018, boeing_osmnx_2017}, engineering design, including architecture, structural optimization, origami design \citep{kakooee_reimagining_2024, ororbia_design_2021, hochuli_inverse_2025, walker_algorithmic_2022, Bleker2024GraphGrammar} as well as computational mesh-generation \citep{Thacher25}. In many of these applications, graphs are not merely abstract combinatorial representations of relational data, but require explicit \emph{planar embeddings} whose structure must be optimized for task-specific objectives under hard structural and geometric constraints. Despite this need, methods for systematically generating goal-directed embedded planar graphs while enforcing hard constraints remain limited in performance or tailored to domain-specific settings~\citep{guo_systematic_2023}.

Classical graph generators~\citep{Erdos_1959_pmd, barbasi_albert, watts_collective_1998} and specialized planar graph generators~\citep{brinkmann_fast_2007, fusy_uniform_2009} provide tractable random graph models, including samplers for well-defined planar graph families. However, they are not generally designed for goal-directed multi-objective optimization nor do they construct valid planar graph embeddings under hard constraints. Deep generative models~\citep{guo_systematic_2023, liu_generative_2023, planar_diffusion_model} can learn graph distributions and support property-conditioned generation~\citep{zhao_controllable_2025}, but they require training data which induces distributional biases and typically handle hard structural or geometric feasibility through soft conditioning, rejection sampling or post-processing. In contrast, autoregressive models, reinforcement learning, and hybrids of both have been used for sequential graph construction under hard validity constraints, especially in molecular design~\citep{shi_graphaf_2020, khemchandani_deepgraphmolgen_2020, zhou_optimization_2019, simm_reinforcement_nodate, xu_reinforced_2020}. Yet many of these methods rely on discrete topological actions or domain-specific validity rules with relatively small graph instances, leaving generic and larger scale goal-directed generation of embedded planar graphs under hard constraints an open challenge.

\begin{figure}[t!]
  \begin{center}
    \centerline{\includegraphics[width=\columnwidth]{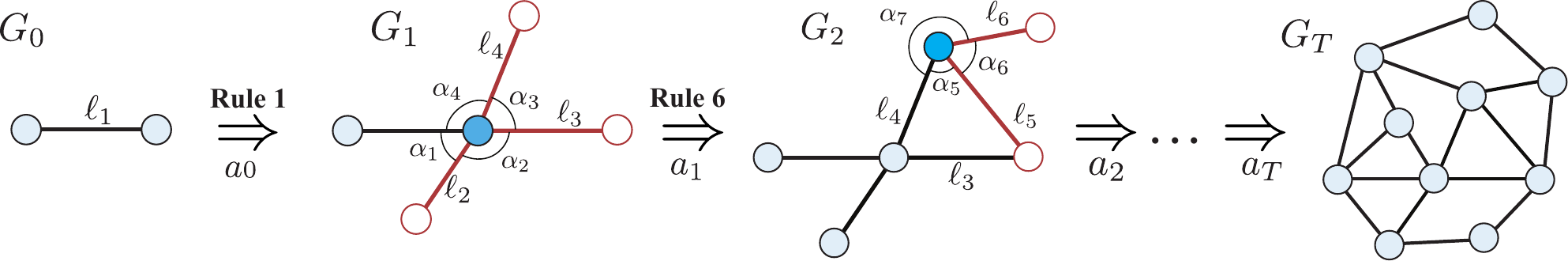}}
    \caption{
        \textbf{Sequential planar graph generation.}
        Starting from an initial graph $G_0$, a policy samples rule-based actions $a_t \sim \pi(a_t \mid G_t)$ from a safe action space, enabling valid graph generation at each step. Actions modify graphs topologically and geometrically based on a predefined graph grammar.} 
    \label{fig:meta_figure}
  \end{center}
  \vskip -0.21in
\end{figure}

Here, we propose a generic dataset-free method for goal-directed embedded planar graph generation across varying graph sizes. The method combines parametric graph grammars with safe reinforcement learning (Figure~\ref{fig:meta_figure}). We formulate the generation process as a Constrained Markov Decision Process (CMDP) whose state and action spaces are induced by the grammar. The grammar defines graph rewriting rules with continuous parameters, while an agent learns to apply these rules sequentially to optimize user-specified multi-objective rewards. Since graph size is controlled directly through sequential rule applications and termination criteria, our method can generate graphs with an exact prescribed number of nodes, a requirement that is difficult to enforce in many distributional graph generators. To improve feasibility during learning, we introduce a differentiable action projection that maps sampled actions toward state-dependent safe sets. The method constructs the planar embedding subject to hard constraints \emph{during} generation, enabling controllable synthesis of planar graph embeddings for settings where multiple optimization objectives jointly matter. We additionally propose a new benchmark suite for constrained goal-directed embedded planar graph generation, that covers a wide variety of graph properties involving geometric, topological and spectral graph metrics.

\section{Preliminaries}
\label{sec:preliminaries}
In this section we introduce parametric graph grammars, as well as their relation to Markov Decision Processes (MDP), to provide the basis for the definition of our method.

\subsection{Graph Grammars}
\label{sec:graph_grammar}
Graph grammars provide a rule-based formalism for the generation
of structured graphs via sequential graph rewriting \citep{kreowski_essentials_2026}. We adopt this formalism to model graph construction in which graph states, admissible rule application sites, and parametric rules are explicitly defined. 
In the following, we formalize the notion of graph states, introduce a parametric graph grammar, and define the graph transition operator that induces state-to-state evolution.

\paragraph{Graph states.}
At each discrete step $t \in \mathbb{N}$ in the rewriting process, we define a graph in its respective state as $    G_t = (V_t, E^d_t, E^b_t, F_t)$,
where $V_t$ is the node set, $E^d_t$ is a set of directed edges forming a
directed acyclic graph (DAG), $E^b_t$ is a set of undirected boundary edges
encoding the planar embedding, $F_t \in \mathbb{R}^{|V_t|\times m}$ assigns
$m$-dimensional feature vectors to nodes, including planar coordinates.
All graphs considered in this work are finite, connected and planar. The primary reason for the DAG formulation is the efficient detection of the \emph{generative frontier} of a graph state $G_t$ which is defined as the set of
leaf nodes $v_t^b \in V_t^b$, called \emph{boundary nodes}, with
\begin{equation}
V_t^b = \{\, v \in V_t \mid \deg^{-}_{E^d_t}(v) \ge 1 \;\wedge\;
\deg^{+}_{E^d_t}(v) = 0 \,\},
\end{equation}
The boundary nodes $v_t^b$ are the only locations at which graph rewriting may be applied.

\paragraph{Parametric graph grammar.}
The graph generation process is directed by a parametric graph grammar $\mathcal{G} = (G_0, \mathcal{P}, \Theta)$, where $G_0$ is an initial graph state, $\mathcal{P}$ is a finite set of rewriting
rules, and $\Theta$ assigns to each rule $p \in \mathcal{P}$ a continuous parameter
space $\Theta(p)$ that spans the geometry of a graph in terms of edge lengths $\ell$ and sector angles $\alpha$.
Each rewriting rule \(p\) is a production \( p: L \to R \) defining a subgraph \( L \) to be matched and replaced from a graph in state \( G_t \) using the subgraph in \( R \). A \emph{match} \( g_t: L \to G_t \) is a graph morphism that identifies a node in the generative frontier such that $g_t(V_L)\subseteq V_t^b$. Since each left-hand side $L$ contains exactly one boundary node,
a match $g_t$ uniquely determines the admissible
rule application site as a boundary node $v_t^b \coloneq g_t(v)$, where $v \in V_L$. The set \( \mathcal{P} \) comprises seven rewriting rules, that we detail in Appendix~\ref{app:rules}.

\paragraph{Graph transition operator.}
At discrete step $t$, a transition $\mathcal{T}$ is determined by selecting
(i) a rule $p_t \in \mathcal{P}$,
(ii) an admissible match $g_t$, and
(iii) a continuous parameter vector $\theta_t \in \Theta(p_t)$.
These choices induce a direct graph transformation,
denoted by $G_t \;\xRightarrow[\mathcal{T}]{}\; G_{t+1}.$  Equivalently, the successor state can be written as $G_{t+1} = \mathcal{T}(G_t; p_t, g_t, \theta_t)$, where $\mathcal{T}$ denotes the \emph{graph transition operator} (see Figure~\ref{fig:meta_figure}).
Operationally, $\mathcal{T}$ introduces new nodes, adds directed edges, updates boundary edges, and assigns planar coordinates to newly created nodes according to the selected rule and its associated parameters. The details regarding the graph transition operator are provided in Appendix~\ref{app:rules}. The graph rewriting process terminates at a finite time $T$ when a termination
criterion is satisfied, such as reaching a prescribed number of nodes in
$G_T$.

\subsection{Constrained Markov Decision Processes}
We briefly review CMDPs, which provide a principled framework for safe sequential decision-making under hard feasibility constraints \citep{altman_constrained_2021}. A CMDP is defined  as  the tuple $(\mathcal{S}, \mathcal{A}, P, r, c, \gamma, \rho_0)$,
where $\mathcal{S}$ is the state space, $\mathcal{A}$ the action space,
$P : \mathcal{S} \times \mathcal{A} \times \mathcal{S} \to [0,1]$ the transition
kernel, $r : \mathcal{S} \times \mathcal{A} \to \mathbb{R}$ the reward function,
$c : \mathcal{S} \times \mathcal{A} \to \{0,1\}$  constraints,
$\gamma \in [0,1)$ the discount factor, and $\rho_0$ the initial-state
distribution. A policy $\pi_\phi: \mathcal{S} \to P(\mathcal{A)}$ is a mapping from states to a probability distribution over actions, where the objective of RL is to find a policy $\pi_\phi(a_t|s_t)$ as the probability of taking action $a_t \in \mathcal{A}$ given the state $s_t \in \mathcal{S}$ at discrete step $t \in \mathbb{N}$ that maximizes the expected discounted cumulative reward as
\begin{equation}
    \label{eq:cmdp_objective}
    \max_\phi \; \mathbb{E}_{\tau \sim \pi_\phi}
    \Big[ \sum_{t=0}^T \gamma^t r(s_t,a_t) \Big]
    \quad \text{s.t.} \quad
    c(s_{t+1}) = 0 \;\; \forall c \in \mathcal{C} \subseteq \{0,1\}.
\end{equation}
% \begin{align}
% \label{eq:cmdp_objective}
% \begin{split}
%             &\max_\phi \; \mathbb{E}_{\tau \sim \pi_\phi} \Big[ \sum_{t=0}^T \gamma^t r(s_t,a_t) \Big] \\
%     s.t. &\quad c(s_{t+1}) = 0 \quad \forall c \in\mathcal{C} \subseteq \{0,1\},
% \end{split}
% \end{align}
where $\tau = (s_0, a_0, s_1,a_1,...)$ denotes a trajectory sampled from the policy $\pi_\phi$ and $\mathcal{C}$ defines the constraint set. The policy is approximated using neural networks with parametrization $\phi$.  In this work, constraints are enforced \emph{instantaneously}: an action $a_t$ is admissible at state $s_t$ only if all $c = 0$ and is otherwise disallowed \citep{liu_policy_2021}. CMDPs of this form can be equivalently viewed as Markov decision processes with state-dependent safe action sets $\mathcal{A}_{\mathrm{safe}}(s_t) \subseteq \mathcal{A}(s_t)$ that respect all constraints in $\mathcal{C}$ for every action taken, i.e.
\begin{equation}
\label{eq:a_safe}
\mathcal{A}_{\mathrm{safe}}(s_t)
=
\big\{ a_t \in \mathcal{A}(s_t)
\;\big|\;
c = 0 \; \forall\, c \in \mathcal{C}(s_{t+1})
\big\}.
\end{equation}

In this setting, the constrained objective in Equation~\ref{eq:cmdp_objective} can be relaxed to an unconstrained formulation, equivalent to standard MDPs.

\subsection{Graph Grammars as CMDPs}
\label{sec:grammars_as_cmdps}
The graph generation  process introduced in
Section~\ref{sec:graph_grammar} implies a CMDP in which each rewriting step corresponds to a state transition.
The CMDP \emph{state space} $\mathcal{S}$ is defined as the set of all admissible
intermediate graph states reachable from the initial graph in $T$ discrete steps under the grammar,
\begin{equation}
\mathcal{S}=
\{\, G \mid G_0 \xRightarrow[\mathcal{T}]{T} G_T \,\},
\end{equation}
while satisfying the constraints in $\mathcal{C}$ at each successive application of $\mathcal{T}$. 
At a given state $G_t$, a safe hybrid action consists of selecting a graph rewriting action $a_t = (g_t, p_t, \theta_t), \; a_t \in \mathcal{A}_{\mathrm{safe}}$ in line with the  operator $\mathcal{T}$. The \textit{transition kernel} is deterministic and directed by the graph
transition operator $\mathcal{T}$,
\begin{equation}
    P(G_{t+1} \mid G_t, a_t)
    =
    \mathbb{I}\{\, G_t \;\xRightarrow[\mathcal{T}]{}\; G_{t+1} \,\}.
\end{equation}
which satisfies the necessary Markov property for CMDPs $P(G_{t+1} \mid G_t, a_t,  \dots, G_0, a_0) = P(G_{t+1} \mid G_t, a_t)$, since transitions depend only on the current state and action.

\paragraph{Summary.}
Under the formulation proposed in this section, the graph generation process using the graph grammar is equivalent to a CMDP subject to hard, instantaneous constraints, where states are represented as graphs and actions are defined as graph rewriting rule applications.
We leverage this formulation in Section~\ref{sec:method} to develop a safe RL algorithm that preserves feasibility throughout generation.

\begin{figure*}[t!]
  %\vskip 0.2in
  \begin{center}
    \centerline{\includegraphics[width=\textwidth]{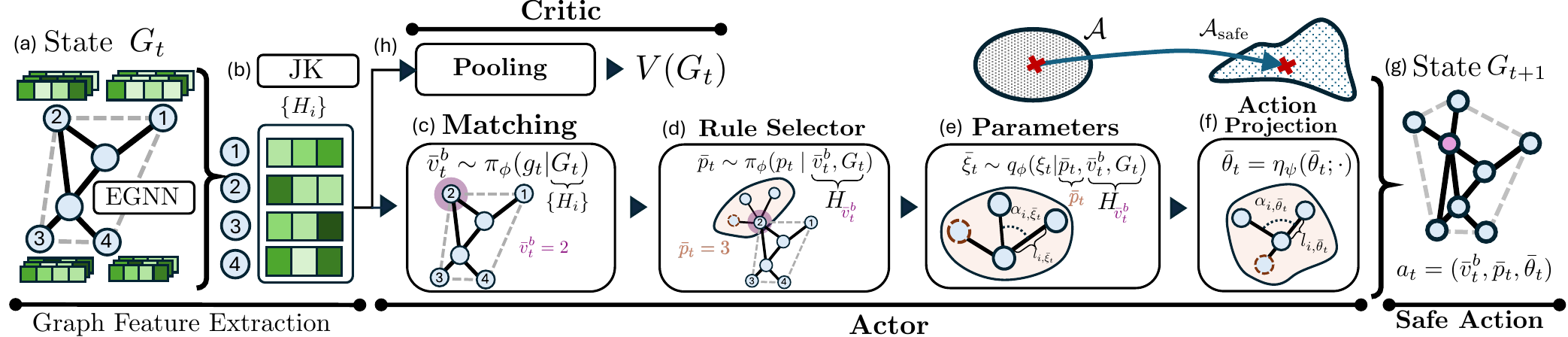 }}
    \caption{Illustration of the proposed Actor-Critic architecture.}
    \label{fig:architecture}
  \end{center}
  \vskip -0.2in
\end{figure*}

\section{Planar Graph Generation}
\label{sec:method}
Building on the CMDP formulation of the parametric graph grammar in Section~\ref{sec:preliminaries}, we formalize planar graph generation as a constrained sequential decision-making problem. The grammar defines  admissible matches and available graph transformations, while a learned policy coordinates the choice of specific match, rule selection and assignments of parameter values to generate graphs optimized for user-specified objectives. Therefore, RL with Proximal Policy Optimization (PPO) \citep{schulman_proximal_2017} is employed to learn a policy that maps the current graph state to a distribution over grammar-based actions, enabling informed decisions over long trajectories $\tau$.
To efficiently account for constraints during learning and inference, we introduce a differentiable action projection mechanism that maps actions onto a learned safe set, enabling stable and sample-efficient policy optimization.

\subsection{General Setup}
The environment state at step $t$ is the graph state $G_t$. The state encodes the full graph
topology, the generative frontier, the planar embedding, and node features. The initial state $G_0$ is defined by two boundary nodes connected by an edge of length 1 (see Figure~\ref{fig:meta_figure}). Based on the preliminaries, the environment dynamics are deterministic for admissible
actions and all stochasticity arises from the policy. We adopt an actor-critic architecture in which both the actor and the
critic operate on a separate graph state encodings using a geometry-aware Graph Neural Network (GNN)-based feature extractor that maps $G_t$ to node- and graph-level feature representations. An overview of the architecture is shown in Figure \ref{fig:architecture} and explained in detail, including training, in Appendix~\ref{app:actor_critic}.

\subsection{Constraint Taxonomy}
\label{sec:contax}
The graph generation process is subject to instantaneous feasibility
constraints, which we explicitly categorize and relate to state-dependent safe
action sets. We define the set of constraints as
\begin{equation}
\mathcal{C}(G_t)
=
\left\{
c_{\ell}(G_t),\\
c_{\alpha}(G_t),\\
c_{\deg}(G_t),\\
c_{\mathrm{pl}}(G_t)
\right\}
%\subset \{0,1\}
\end{equation}
\begin{comment}
\begin{equation}
\mathcal{C}(G_t)
=
\left\{
\begin{aligned}
c_{\ell}(G_t),\\
c_{\alpha}(G_t),\\
c_{\deg}(G_t),\\
c_{\mathrm{pl}}(G_t)
\end{aligned}
\right\}
\subset \{0,1\}.
\end{equation}
\end{comment}
where the first constraint $c_\ell(G_t) = 0$ if $\ell(e) \in [\ell_{\min}, \ell_{\max} ] \; \forall e \in E_t^d \cup E_t^b$, i.e. all edge lengths in the graph are within bounds, otherwise $1$. The second constraint $c_{\alpha}(G_t) = 0$ if all $\alpha_i (v) > \alpha_{\min} \; \forall v \in V_t$ representing minimum bounds on all sector angles $\alpha_i$ between consecutive incident edges around all nodes $v$ of $G_t$, otherwise 1. The third constraint $c_{\deg}(G_t)=0$ if $\deg(v) \leq \deg_{\max} \forall v\in V_t$, otherwise 1, representing maximum degree bounds on nodes. The last constraint $c_\mathrm{pl}(G_t)$ checks whether the graph is drawn as a valid planar embedding, i.e. that no edge intersections occur, returns $0$ if so, otherwise $1$. 
Hence, a graph transition $\mathcal{T}$ is only admissible if $c = 0 \; \forall\, c \in \mathcal{C}(\mathcal{T}(G_t;a_t))$.

We further decompose the constraint set as $\mathcal{C} = \mathcal{C}_T \cup \mathcal{C}_G$ into topological and geometric constraints, respectively.
Topological constraints $\mathcal{C}_T$ enforce the structural admissibility of an action and cannot be resolved by changing the values of continuous parameters $\theta_t$ after a rule $p_t$ and a match $g_t$ are determined. Such violations are intrinsic to the graph topology generated by a rewriting rule and therefore irreversible in our set-up. Geometric constraints $\mathcal{C}_G$, in contrast, correspond to violations that can, in principle, be resolved by finding the appropriate values of $\theta_t$ given a rule $p_t$ and a match $g_t$. These violations occur due to boundary edge-length violations in $E_t^b$ or edge intersections. In practice, violations of $\mathcal{C}_T$ occur infrequently (see Appendix~\ref{app:action_projection_results}) and thus can be simply handled by rejection sampling over the discrete components of the policy. In contrast,
violations of $\mathcal{C}_G$ occur at a high rate, particularly edge
intersections, and naive rejection sampling would lead to severe sampling
inefficiency. This motivates the introduction of an explicit action projection
mechanism that maps nominal parametrization of rules directly to a safe set $\mathcal{A}(G_t) \to \mathcal{A}_{\mathrm{safe}}(G_t)$, see Section~\ref{sec:agent_design}.

\subsection{State, Feature Extraction and Critic}
\label{sec:feature_extraction}
\paragraph{Graph observation.}
At each discrete step, the environment returns the current graph state $G_t$ with the
corresponding node feature matrix $F_t \in \mathbb{R}^{|V_t|\times m}$. Node features
consist of (i) a binary boundary node indicator $b_i$ for each node, (ii) the integer node degree
$d_i$, and (iii) a degree-invariant encoding of local sector angles using
spherical harmonics, similar to
\citep{liu_spherical_2022, gasteiger_directional_2022}. These are computed using
the first $K=4$ Fourier harmonics as
\begin{equation}
    h_i = (\bar c_{i,1}, \bar s_{i,1}, \dots, \bar c_{i,K}, \bar s_{i,K}) \in \mathbb{R}^8,
\end{equation}
with
$\bar c_{i,k} = \frac{1}{d_i}\sum_{m=1}^{d_i}\cos(k\alpha_{i,m})$ and
$\bar s_{i,k} = \frac{1}{d_i}\sum_{m=1}^{d_i}\sin(k\alpha_{i,m})$. Finally, we add the absolute two-dimensional node coordinates to the node attributes as $y_i$, and stack $x_i = (b_i, d_i, h_i, y_i)$  into $F_t$.

\paragraph{Graph feature extraction and Critic.}
To enable decision-making over graph states, the graph observation is encoded using 
E(2)-Equivariant Graph Neural Networks (EGNN) \citep{satorras_en_2022} that jointly processes graph structure and planar embedding, where we extract the invariant features only. Prior to the EGNN, node attributes $(b_i, d_i, h_i)$ are lifted to a higher dimension via a small Multilayer Perceptron (MLP). To accommodate growing graph sizes, we adopt the Jumping Knowledge (JK) mechanism \citep{xu_representation_2018} with attentional aggregation scaled by graph size. The resulting node embeddings are combined with  a global graph representation obtained via attention pooling \citep{li_graph_2019} and mean pooling. These are concatenated into a set of node-wise features $\{H_i\}_{i=1}^{|V_t|}$, which serve as input to the downstream agent. Further details on the  feature extractor architecture are given in Appendix~\ref{app:feature_extractor}.

The critic employs an equivalent encoder architecture which is trained separately. In this setting, only the global graph representation is used, augmented with a learned embedding of the graph-size  $\log(1 + |V_t|)$, and passed through a small MLP to predict the state value $V(G_t)$.

\subsection{Agent Design}
\label{sec:agent_design}
We now define the hybrid agent design and the associated conditional policy,
aligned with the semantics of the graph transition operator
$\mathcal{T}(G_t; g_t, p_t, \theta_t)$. Architectural details on the policy networks are outlined in Appendix~\ref{app:policy_networks}.

\paragraph{Conditional hybrid policy.}
At each discrete step, the agent applies a graph rewriting action
\begin{equation}
a_t = (g_t, p_t, \theta_t)
\;\in\;
\mathcal{A}(G_t)
\subseteq
V_t^b\times \mathcal{P}\times \Theta(p_t),
\end{equation} 
where the policy is structured autoregressively. First, the agent samples a boundary node
according to $\pi_\phi(g_t \mid G_t)$, implemented as a categorical distribution
over boundary nodes masked by $b_i$ and approximated using an MLP with input $\{H_i\}$. This yields
$\bar v_t^b \sim \pi_\phi(g_t \mid G_t)$.

Conditioned on $\bar v_t^b$, the agent samples a rewriting rule from the categorical distribution  $\bar p_t \sim \pi_\phi(p_t \mid \bar v_t^b, G_t)$, which is another MLP with input $H_{\bar v_t^b}$, i.e. the node embedding for the selected rule application site. Since violations of $\mathcal{C}_T$ are rare at this stage (see Figure~\ref{fig:action_proj_summary}) but cannot be ruled out, we apply rejection sampling over $p_t$, advancing $t\!\to\!t+1$ with zero reward and no state transition, while signaling rejection to PPO through the discount $\gamma<1$. With this approach, we limit $(\bar v_t^b, \bar p_t$) to the safe subset of $V_t^b\times \mathcal{P}$.

Finally, conditioned on $(\bar v_t^b, \bar p_t)$, the agent samples continuous
parameters from $\bar \theta_t \sim \pi_\phi(\theta_t \mid \bar v_t^b,\bar p_t, G_t)$. This
distribution is obtained by transforming a nominal Gaussian
$q_\phi(\xi_t \mid \cdot) = \mathcal{N}(\mu_\phi, \sigma_\phi^2)$ via a push-forward $\tanh$-affine map. The Gaussian mean is predicted by an MLP with input $H_{\bar v_t^b}$ and $\bar p_t$, while the
standard deviation is a global trainable parameter. For each rewriting rule, the dimensionality of the Gaussian dim$(\Theta(p))$ is different, so we apply masking to retrieve rule-dependent distribution dimensions.  The $\tanh$-affine map projection serves two purposes: it enforces the boundedness
$\theta_t \in (-1,1)$, and it enables the approximate projection of nominal samples $\xi_t$ into the geometric safe set $\mathcal{A}_{\mathrm{safe}}$ induced by $\mathcal{C}_G$. We refer to this mechanism as \emph{action projection}. 

\paragraph{Action Projection.}
Nominal parametrizations $\xi_t$ with infinite support are projected using a
diffeomorphism $\eta_\psi: \mathbb{R}^{\dim(\Theta(p_t))} \to (-1,1)^{\dim(\Theta(p_t))}$,
\begin{equation}
\label{eq:action_projection}
\theta_t
=\tanh\!\big(
s_\psi(G_t,g_t,  p_t) \odot \xi_t+t_\psi(G_t,g_t,  p_t)\big),
\end{equation}
where $s_\psi$ and $t_\psi$ are learned, state-dependent, scaling and shift
functions with parameters $\psi$. This construction is equivalent to a normalizing flow \citep{rezende_variational_2016} with a single flow step followed by a $\tanh$ squashing. We found this one-step flow to be sufficient empirically. We further explain the construction of $s_\psi$ and $t_\psi$ in Section~\ref{sec:safe_action_projection_learning}. Applying the action projection to the policy's distribution as $\pi_\phi(\theta_t|g_t, p_t, G_t)=\eta_{\psi_\#} q_\phi(\xi_t \mid g_t, p_t, G_t)$ results in the push-forward measure
\begin{align}
    \pi_\phi(\theta_t \mid \cdot)
= 
q_\phi(\xi_t\mid  \cdot)\,
\bigl|\det J_\psi(g_t, p_t, G_t)\bigr|^{-1}, \quad \det J_\psi(g_t, p_t, G_t)
= \prod_i
\Big(s_{\psi,i}\, (1-\theta_{t,i}^2)\Big)
\end{align}
where the expression follows directly from the change-of-variables correction for densities and $J_\psi$ denotes the  Jacobian $\frac{\partial\eta_\psi}{\partial\xi_t}(\xi_t)$, which is diagonal since the projector acts element-wise. %$J_\psi$ is diagonal and the determinant computed as
%where $\det J_\psi(g_t, p_t, G_t) \neq 0 \; \forall \xi_t$ if %$s_{\psi,i}\neq0$.

\paragraph{Joint policy.}
The resulting joint policy $\pi_\phi(a_t\mid G_t)$ after action projection factorizes as
\begin{align}
\pi_\phi(a_t \mid G_t)
&= \;\pi_\phi(g_t\mid G_t)\,\pi_\phi(p_t\mid g_t, G_t)\,\pi_\phi(\theta_t\mid g_t, p_t, G_t) \\
= &\;\pi_\phi(g_t\mid G_t)\,\pi_\phi(p_t\mid g_t, G_t)\,q_\phi(\xi_t \mid g_t, p_t, G_t)\, \notag\bigl|\det J_\psi(g_t, p_t, G_t)\bigr|^{-1},
\end{align}
which is then used in PPO for policy learning. 
%Note that in this formulation, we do not optimize a density directly in $\theta_t$, which would be intractable. Instead, the joint distribution uses the change-of-variables correction induced by $J_\psi$ over the nominal Gaussian in $\xi_t$, while the parameters $\psi$ of the projector are frozen during a PPO update. 
To encourage exploration under PPO via an entropy bonus, we compute the entropy
using the same factorization:
\begin{equation}
\label{eq:entropy}
\mathcal{H}[\pi_\phi]
= w_g\,\mathcal H[g_t \mid G_t]
+ w_p\,\mathbb E_{g_t\sim \pi_\phi(\cdot)}\!\big[\mathcal H[p_t\mid g_t,G_t]\big]
+ w_\xi\,\mathbb E_{g_t,p_t \sim \pi_\phi(\cdot)}\!\big[\mathcal H[\xi_t\mid p_t, g_t, G_t]\big]
\end{equation}
We omit the Jacobian term due to missing closed-form entropy and because its Monte Carlo estimation can be noisy, and compute the continuous entropy over the nominal distribution $q_\phi(\xi_t \mid \cdot)$. Categorical and continuous components are weighted by $w_{(\cdot)}$ (set to $0.9$ and $0.1$, respectively). Policy MLPs are initialized with small final-layer weights and zero bias to encourage near-uniform initial exploration.

\subsection{Action Projection Learning}
\label{sec:safe_action_projection_learning}
The projection in Equation~\ref{eq:action_projection} specifies \emph{how} nominal samples
$\xi_t$ are mapped into bounded parameters $\theta_t \in (-1,1)$, but it does not by itself
specify \emph{where} to project a given $\xi_t$ in order to satisfy geometric
constraints. To obtain supervision for learning a projector that targets the
admissible safe set $\mathcal{A_\mathrm{safe}}$, we run an efficient implementation of Covariance Matrix Adaptation - Evolutionary Strategy (CMA-ES) \citep{hansen_completely_2001} inside the environment during graph rewriting on the continuous parameters $\theta_t'$, with the projections' initial guess $\bar \theta_t$, to resolve potential violations of $\mathcal{C}_G$ using the formulation
\begin{equation}
\label{eq:cmaes}
\theta_t^\star = \arg\min_{\theta_t'}\; \|\theta_t'- \bar\theta_{t}\|_2 \qquad
\mathrm{s.t.} \quad c(\mathcal{T}(G_t;g_t,p_t,\theta_t)) = 0 \quad \forall c \in \mathcal{C}_G,
\end{equation}
which follows the paradigm of safety-layer approaches in safe RL \citep{dalal_safe_2018, liu_policy_2021}. The key
idea is to find a proximal $\theta_t^\star$ to the projection's previous guess $\bar\theta_{t}$, that satisfies the constraints, implemented via an explicit penalty-based formulation (see Equation~\ref{eq:cmaes_penalty_objective} in the Appendix). The resulting candidate $\theta_t^\star$ is accepted only if all $c = 0$, otherwise we fall back to the rejection sampling scheme. Empirically, CMA-ES succeeds in $70$--$90\%$ of cases to find such a $\theta_t^\star$. If the projection's guess $\bar \theta_t$ is already admissible, we set $\theta_t^\star=\bar \theta_t$. Then, using the corrected $\theta_t^\star$ obtained from CMA-ES and nominal samples $\bar \xi_t$, we assemble an off-policy replay buffer to train the projection $\eta_\psi: \bar \xi_t \mapsto \bar \theta_t^\star$ in a supervised setting alongside PPO, which continuously refines the projection through CMA-ES correction. Ultimately, the action projector converges toward the admissible safe-set projection, CMA-ES corrections become in-frequent and act only as a fallback, as shown in Appendix~\ref{app:action_projection_results}, where we also provide computational overhead insights of CMA-ES. The projector architecture uses an equivalent graph feature extractor (as
introduced in Section~\ref{sec:feature_extraction}) for $G_t$. Given the node embedding at a selected $\bar v_t^b$ and a rule embedding for the selected $\bar p_t$, we use two small MLPs to compute the scale $s_\psi$ and shift $t_\psi$, with $s_\psi>0$. The projector is trained synchronously with respect to the policy parameters. Details about architecture and training are given in Appendix~\ref{app:action_projector}. 

\subsection{Reward Design}
\label{sec:reward}
The reward function measures how closely the current graph satisfies a set of user-defined objectives, expressed through selected metrics (Section~\ref{sec:Experiments}). Let $\mathcal{M}$ denote this set. For each metric $m \in \mathcal{M}$, we compute a symmetric relative deviation (lower is better) where $\kappa^\star$ and $\kappa$ denote the target and current values of metric $m$, respectively, and $\epsilon > 0$ ensures numerical stability. The multiobjective reward $r$, which encourages solutions that jointly satisfy all metrics, is then defined as
\begin{equation}
\label{eq:reward}
 r = 1 - \frac{1}{\lvert \mathcal{M} \rvert}\sum_{m \in \mathcal{M}} d_m, \qquad 
    d_m = \frac{\lvert \kappa^\star - \kappa \rvert}{\lvert \kappa^\star \rvert + \lvert \kappa \rvert + \epsilon} \in [0,1].
\end{equation}

\section{Related Works}
\label{sec:related_work}
Classical graph models such as Erd\H{o}s-R\'enyi or Barab\'asi-Albert graphs~\citep{Erdos_1959_pmd, barbasi_albert}, as well as specialized planar graph generators~\citep{brinkmann_fast_2007, fusy_uniform_2009}, provide useful baselines and, in some cases, planarity guarantees at the abstract graph level. However, they are not designed for goal-directed multi-objective optimization or for constructing task-optimized planar embeddings under hard geometric constraints. Deep graph generative models can learn graph distributions with high distributional fidelity~\citep{guo_systematic_2023, liu_generative_2023, planar_diffusion_model}, and conditional or guided variants can bias samples toward desired properties~\citep{kunegis_guided_2023, bergmeister_efficient_2024}. Models such as DiGress and GDSS with Prodigy guidance generate abstract graph structures and can be adapted to multi-objective and single-objective property-conditioned generation, respectively~\citep{vignac_digress_2023, jo2022scorebasedgenerativemodelinggraphs, pmlr-v235-sharma24b}. However, enforcing exact graph size, planarity, and hard structural constraints through conditioning remains difficult, with validity often relying on rejection sampling, post-processing, or external guidance~\citep{zhao_controllable_2025}. TreeDiff \cite{zhao_controllable_2025} extends pretrained graph diffusion models with verifier-guided tree search at inference time but adapting it to embedded planar graphs requires task-specific modifications to the diffusion backbone, intermediate-state verifier, and search procedure. Geometry-aware models such as NeVAE and GeoDiff primarily target molecular domains and rely on chemistry-specific validity assumptions~\citep{samanta2019nevaedeepgenerativemodel, xu2022geodiffgeometricdiffusionmodel}. MuDiff more directly couples discrete structure and continuous coordinates~\citep{hua2024mudiffunifieddiffusioncomplete}, but feasibility is still learned implicitly from data rather than enforced as a construction constraint.

Reinforcement learning provides a natural framework for goal-directed graph construction through sequential decisions~\citep{darvariu_goal-directed_2021, nie_reinforcement_2023}. It has been widely used in molecular graph generation, where policies add atoms and bonds under chemistry-specific validity~\citep{you_graph_2018, zhou_optimization_2019, khemchandani_deepgraphmolgen_2020, shi_graphaf_2020}.  Grammar- and rule-based RL has been applied to molecular optimization, where policies select productions from molecular grammars to preserve chemical validity \cite{xu_reinforced_2020, kraev_grammars_2018}, to structural design, where grammar actions preserve domain-specific structural constraints \cite{ororbia_design_2021, Bleker2024GraphGrammar}. Outside generative settings, many graph-based RL methods focus on tasks such as adversarial attacks, knowledge-graph completion, or relational reasoning rather than generic graph generation~\citep{nie_reinforcement_2023}. It has also been applied to computation-graph optimization, where RL selects rules for neural-network subgraph rewriting \cite{he_x-rlflow_nodate}. These works demonstrate the value of structured action spaces. However, they mainly address discrete and structural domain-specific construction.   Our distinguishing setting is the goal-directed generation of planar embedded graphs using hybrid discrete-continuous action spaces: the policy jointly selects a graph match, a rewriting rule, and safe continuous geometric parameters, thereby coupling discrete topological graph rewriting with continuous embedding optimization under state-dependent topological and geometric constraints. 
%Our method extends grammar-based RL to embedded planar graphs by combining discrete rewriting decisions with continuous geometric parametrization and explicit feasibility checks during construction.

\section{Experiments}
\label{sec:Experiments}
We evaluate our method in two stages. First, we introduce four multi-objective benchmarks to assess the method's performance under varying geometric, topological and spectral objective as well as different constraint tightness. Second, we ablate the proposed action projection on a single-objective graph generation task and perform a scalability study. All results are averaged over 5 seeds with 1-sigma standard deviations reported for the rewards achieved by our method. Training was performed on an NVIDIA RTX PRO 6000 GPU, while the environment ran on an AMD 16-Core CPU; runtimes ranged from 10 minutes to 2 hours depending on benchmark complexity.

\subsection{Benchmarks}
\label{sec:benchmark_experiments}

To the best of our knowledge, no standardized benchmarks have been established for constrained, goal-directed generation of planar embedded graphs. Accordingly, the proposed benchmark suite comprises four heterogeneous tasks combining geometric, topological, and spectral metrics for constrained multi-objective optimization and varying graph sizes $|G|$, where all metrics used are defined in detail in Appendix~\ref{app:metrics}. Table~\ref{tab:benchmarks} reports benchmark results and compares our method against the three best baselines out of all 10 reported in Appendix~\ref{app:extended_results}. This includes the best results from the \emph{House of Graphs} (HOG) dataset~\citep{brinkmann_house_2013}, which contains more than $4{,}000$ planar graphs with graph size up to 250. Since HOG graphs are not embedded, we compute embeddings using a force-directed spring layout~\citep{spring_layout}. Further, we compare against three diffusion-based generators; DiGress, GDSS, and MuDiff~\citep{vignac_digress_2023, jo2022scorebasedgenerativemodelinggraphs, hua2024mudiffunifieddiffusioncomplete} on the same dataset and report the strongest contender for each benchmark. These models were selected as they allow for conditional generation, in the case of DiGress and MuDiff, while GDSS can be guided towards graph size using Prodigy~\citep{pmlr-v235-sharma24b}. The classical generator baselines used are described in Appendix~\ref{app:graph_gen_benchmark}. For all classical baselines, we apply ranking-based selection to retrieve graphs satisfying the most constraints and, among them, maximize the reward. For all deep generative baselines, we evaluate from $1000$ generated candidates. After retaining the largest connected component and filtering out non-planar graphs, candidates are ranked first by proximity to the target graph size and then by benchmark reward. The remaining geometric and structural constraint statistics are evaluated and reported, but they are not used for candidate selection. This alternative protocol for deep generative models is motivated by the fact that we empirically found diffusion models to rarely satisfy the constraint set, and prioritizing constraint satisfaction would substantially reduce the reward performance of these models. We consequently exclude the constraint satisfaction criterion from the ranking procedure for diffusion baselines to enable a fairer comparison of reward metrics. More details on the evaluation protocol are available in Appendix~\ref{app:graph-selection-proc}. For benchmark 1, 2, 4, we enforce a tight constraint set $\mathcal{C}_2$ with $\ell \in [0.75,2.5]$, $\alpha_{\min}=20^{\circ}$, and $\deg_{\max}=8$. For benchmark 3, we use the relaxed constraint set $\mathcal{C}_1$ with $\ell \in [0.1,3.5]$,  $\alpha_{\min}=1^{\circ}$ and $\deg_{\max}=15$ to allow for fairer comparison against unconstrained baselines.

\begin{table}[t]
\centering
\caption{Benchmark results across four planar graph generation tasks. Each graph corresponds to the result obtained by our method. The table shows averaged multi-objective reward, the individual metrics, graph sizes and number of satisfied constraints for each benchmark. Evaluations for all 10 baselines are reported in Appendix~\ref{app:extended_results}, complemented with per-baseline graph visualizations. Relative metric deviations from the targets are additionally provided in Appendix Table~\ref{tab:all-benchmark-deviations}.}
\vspace{0.4em}
\label{tab:benchmarks}
\setlength{\tabcolsep}{3.0pt}
\renewcommand{\arraystretch}{0.93}
\begin{tabular}{lclccccccc}
\toprule
Benchmark
& Graph
& Method
& Reward
& \multicolumn{4}{c}{Metrics}
& \multicolumn{2}{c}{Constr.} \\
\cmidrule(lr){1-1}
\cmidrule(lr){2-2}
\cmidrule(lr){3-3}
\cmidrule(lr){4-4}
\cmidrule(lr){5-8}
\cmidrule(lr){9-10}

\multirow{6}{*}{\shortstack[l]{1. Resilient\\ Power Grid}}
& \multirow{6}{*}{\includegraphics[width=0.165\linewidth]{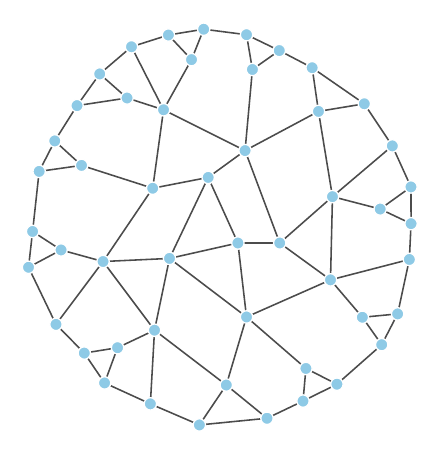}}
&
& $r$
& $r_{\mathrm{iso}}$
& $c_{\mathrm{gini}}$
& $c_{\mathrm{clust}}$
& $\lambda_2$
& $|G|$
& $\mathcal{C}_2$ \\
\cmidrule(lr){3-10}
&
& Target   & 1.0 & 1.0 & 0.1 & 0.2 & 1.0 & 50 & \cmark  \\
\cmidrule(lr){3-10}
&
& \cellcolor{oursblue}\textbf{Ours}
& \cellcolor{oursblue}\textbf{0.76}\scriptsize{$\pm$ 0.004}
& \cellcolor{oursblue}\textbf{0.96}
& \cellcolor{oursblue}\textbf{0.09}
& \cellcolor{oursblue}\textbf{0.2}
& \cellcolor{oursblue}0.05
& \cellcolor{oursblue}\textbf{50}
& \cellcolor{oursblue}\cmark ($\sfrac{4}{4}$)  \\
&
& HOG      & 0.38 &  0.84  & 0.03 & 0.0 & 0.08 & 47 & \ding{55} ($\sfrac{3}{4}$)  \\
&
& Delaunay & 0.69 & 0.93 & 0.18 & 0.22 & \textbf{0.09}& \textbf{50} &\ding{55}  ($\sfrac{2}{4}$)  \\
&
& GDSS  & 0.56 & 0.50 & 0.2 & 0.16 & 0.01 & 43  & \ding{55} ($\sfrac{2}{4}$)  \\

\specialrule{0.08em}{0.45em}{0.25em}

\multirow{6}{*}{\shortstack[l]{2. Urban\\ Road Grid}}
& \multirow{6}{*}{\includegraphics[width=0.165\linewidth]{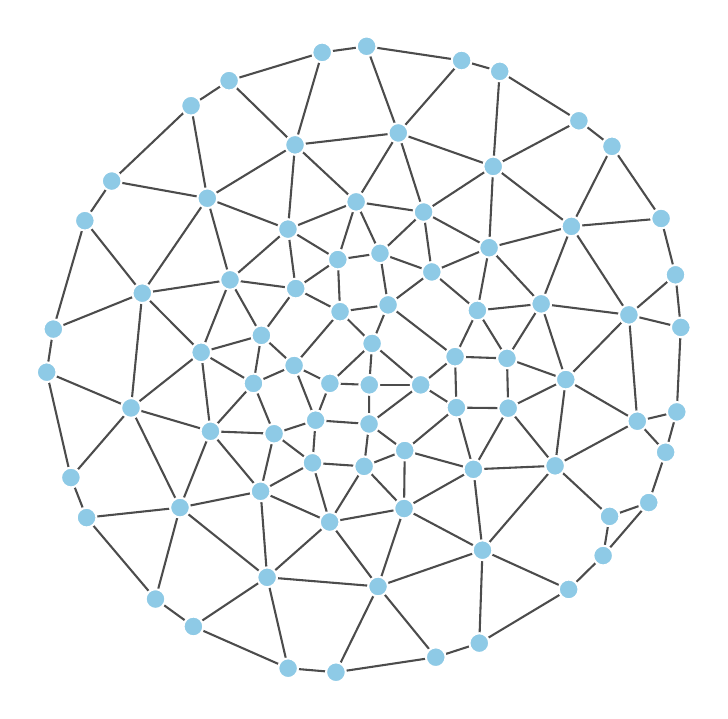}}
&
& $r$
& $r_{\mathrm{iso}}$
& $S_l$
& $\mathrm{AR}$
& $c_{\mathrm{gini}}$
& $|G|$
& $\mathcal{C}_2$ \\
\cmidrule(lr){3-10}
&
& Target   & 1.0 & 1.0 & 0.7 & 0.8 & 0.2 & 80 & \cmark \\
\cmidrule(lr){3-10}
&
& \cellcolor{oursblue}\textbf{Ours}
& \cellcolor{oursblue}\textbf{0.94}\scriptsize{$\pm$ 0.01}
& \cellcolor{oursblue}\textbf{0.95}
& \cellcolor{oursblue}0.82
& \cellcolor{oursblue}\textbf{0.76}
& \cellcolor{oursblue}0.16
& \cellcolor{oursblue}\textbf{80}
& \cellcolor{oursblue}\cmark ($\sfrac{4}{4}$)\\
&
& HOG  & 0.78 & 0.4 & 0.83 & 0.49 & 0.16 & \textbf{80} & \ding{55} ($\sfrac{1}{4}$)\\
&
&  Delaunay  & 0.91 & \textbf{0.95} & \textbf{0.76 }& 0.45 & \textbf{0.2} & \textbf{80} & \ding{55} ($\sfrac{2}{4}$)\\
&
& MuDiff & 0.53 & 0.34 & 0.01 & 0.46 & 0.25 & 56 & \ding{55} ($\sfrac{1}{4}$) \\

\specialrule{0.08em}{0.45em}{0.25em}

\multirow{6}{*}{\shortstack[l]{3. Strongly\\ Connected\\ Mesh }}
& \multirow{6}{*}{\includegraphics[width=0.165\linewidth]{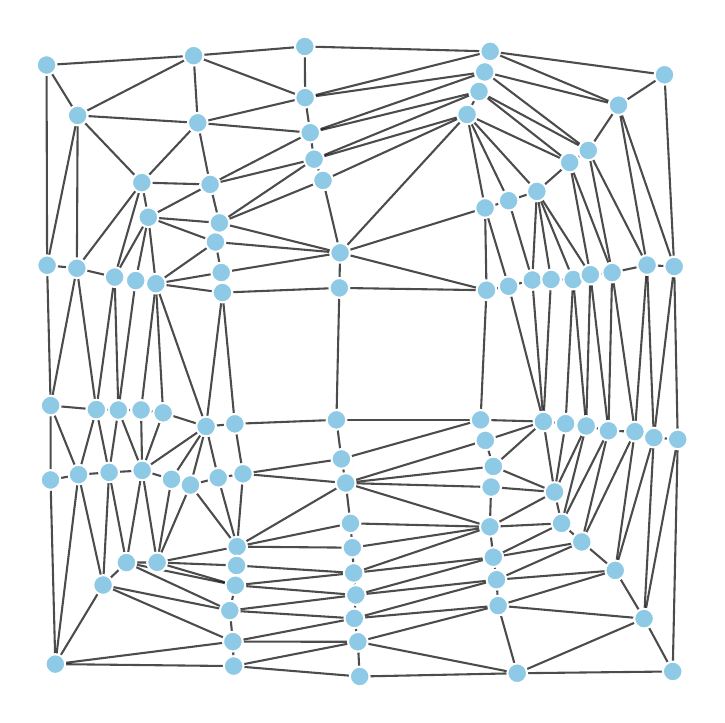}}
&
& $r$
& $r_{\mathrm{square}}$
& $\lambda_2$
& --
& --
& $|G|$
& $\mathcal{C}_1$ \\
\cmidrule(lr){3-10}
&
& Target   & 1.0 & 1.0 & 0.3 & -- & -- & 100 & \cmark \\
\cmidrule(lr){3-10}
&
& \cellcolor{oursblue}\textbf{Ours}
& \cellcolor{oursblue}\textbf{0.65}\scriptsize{$\pm$ 0.01}
& \cellcolor{oursblue}0.89
& \cellcolor{oursblue}\textbf{0.06}
& \cellcolor{oursblue}--
& \cellcolor{oursblue}--
& \cellcolor{oursblue}\textbf{100}
& \cellcolor{oursblue}\cmark ($\sfrac{4}{4}$) \\
&
& HOG & 0.36 & 0.54 & 0.00 & -- & -- & 101 & \ding{55}($\sfrac{3}{4}$)\\
&
& Lattice  & 0.59 & \textbf{1.0} & 0.03 & -- & -- & \textbf{100} & \cmark ($\sfrac{4}{4}$)\\
&
& GDSS   & 0.39 & 0.55 & 0.01 & -- & -- & 72 & \ding{55}($\sfrac{3}{4}$)\\

\specialrule{0.08em}{0.45em}{0.25em}

\multirow{6}{*}{\shortstack[l]{4. Graph with\\ 40 Triangles}}
& \multirow{6}{*}{\includegraphics[width=0.165\linewidth]{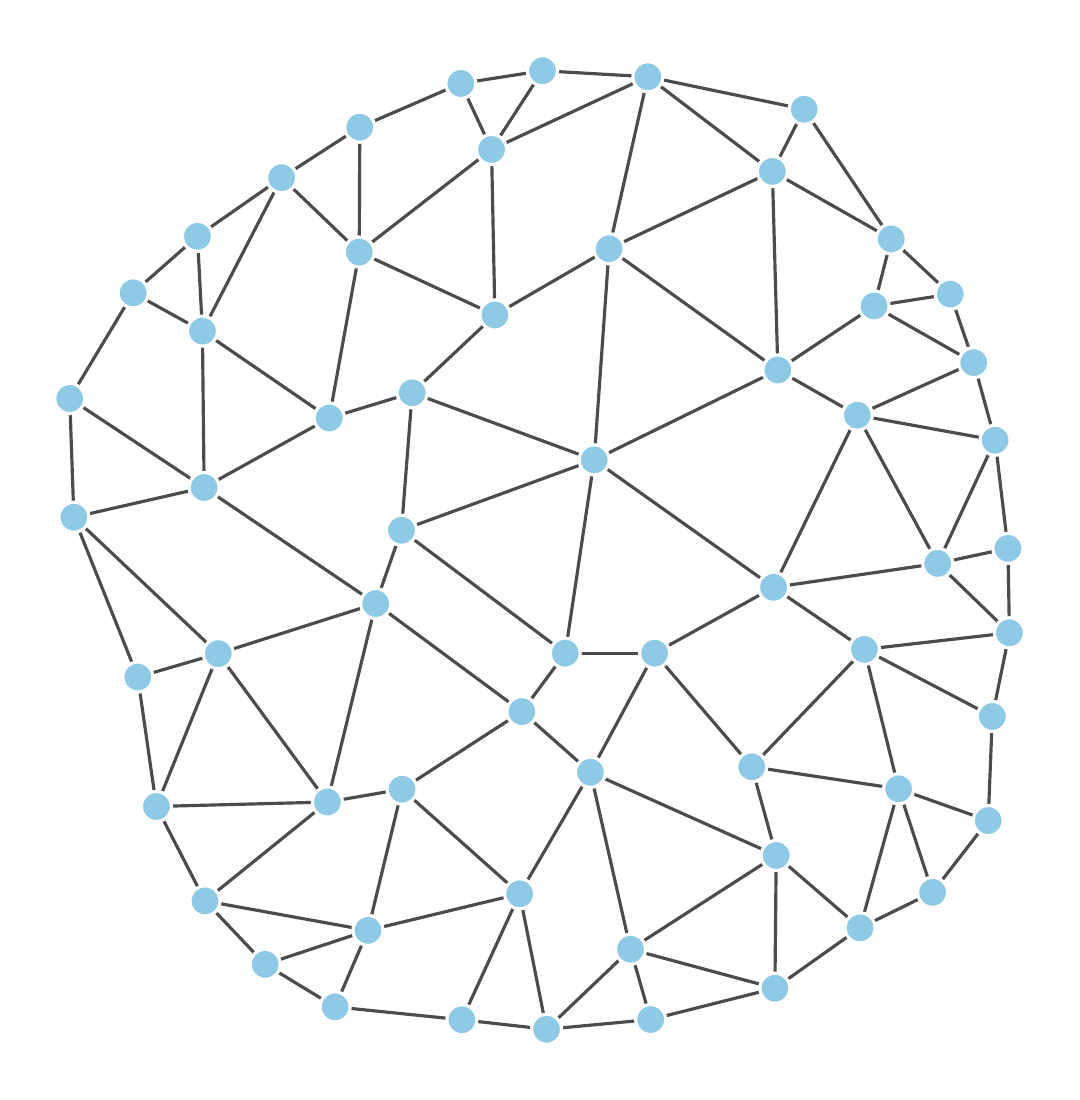}}
&
& $r$
& $r_{\mathrm{iso}}$
& $T$
& --
& --
& $|G|$
& $\mathcal{C}_2$ \\
\cmidrule(lr){3-10}
&
& Target   & 1.0 & 1.0 & 40 & -- & -- & 60 & \cmark \\
\cmidrule(lr){3-10}
&
& \cellcolor{oursblue}\textbf{Ours} 
& \cellcolor{oursblue}\textbf{0.98}\scriptsize{$\pm$ 0.02}
& \cellcolor{oursblue}\textbf{0.92}
& \cellcolor{oursblue}\textbf{39.91}
& \cellcolor{oursblue}--
& \cellcolor{oursblue}--
& \cellcolor{oursblue}\textbf{60}
& \cellcolor{oursblue}\cmark ($\sfrac{4}{4}$) \\
&
& HOG & 0.86 & 0.89 & 64 & -- & -- & \textbf{60} & \cmark  ($\sfrac{4}{4}$) \\
&
& Boltzmann  & 0.88 & 0.63 & 38 & -- & -- & \textbf{60} & \ding{55} ($\sfrac{0}{4}$)  \\
&
& DiGress   & 0.57 & 0.42 & 15 & -- & -- & 10 & \ding{55} ($\sfrac{2}{4}$)  \\

\bottomrule
\end{tabular}
\end{table}

\paragraph{Resilient power grid in circular hull.}
This benchmark generates a 50-node power-grid-like graph inside a circular hull, measured by the isoperimetric ratio $(r_\mathrm{iso}=1.0)$. The objective combines shape matching, low degree inequality $(c_\mathrm{gini}=0.1)$, moderate clustering $(c_\mathrm{clust}=0.2)$, and normalized spectral gap $(\lambda_2 = 1.0)$ measuring overall connectivity. Here, the target $\lambda_2=1.0$ is idealistic; planar, bounded-degree, geometrically constrained graphs have much lower attainable spectral gaps, so this term acts primarily as a connectivity pressure. Results show that our method  balances these competing metrics best, while satisfying all constraints. Note here, that Delaunay triangulation, representing a maximally connected planar embedded graph, achieves highest spectral gap, which is this baselines natural strength, but does so at the cost of clustering and constraint satisfaction. 

\paragraph{Urban road grid in circular hull.}
This benchmark synthesizes an 80-node road-network-like graph in a circular hull. The objective combines shape matching $(r_\mathrm{iso})$, edge-length regularity $(S_l=0.7)$, angular resolution $(\mathrm{AR}=0.8)$, and low degree inequality $(c_\mathrm{gini}=0.2)$. Although Delaunay triangulation shows a better match on two individual metrics, it does so by strongly underperforming on angular resolution and falls behind in total reward while also failing to satisfy 2/4 constraints.

\paragraph{Strongly connected mesh in square hull.}
This benchmark generates a 100-node mesh in a square hull with high connectivity, combining a squareness objective $(r_\mathrm{square}=1.0)$ with a spectral-gap objective $\lambda_2$. Similar to benchmark 1, the goal of $\lambda_2 =0.3$ is chosen arbitrarily as a maximization pressure on connectivity. Lattice graphs naturally achieve perfect squareness, but have lower spectral gap. Our method provides the best overall trade-off while satisfying all constraints.

\paragraph{Graph with 40 triangles in circular hull.}
This benchmark targets a 60-node mesh in a circular hull with exactly 40 triangles. Unlike standard triangulation methods, the task requires a specific combinatorial face. Our method closely matches the triangle target on average and satisfies all constraints, whereas HOG substantially overshoots the triangle count.

\subsection{Ablation and Scalability Study}
\begin{figure}[t]
    \centering

    \begin{minipage}[t]{0.43\linewidth}
        \centering
        \includegraphics[width=0.75\linewidth]{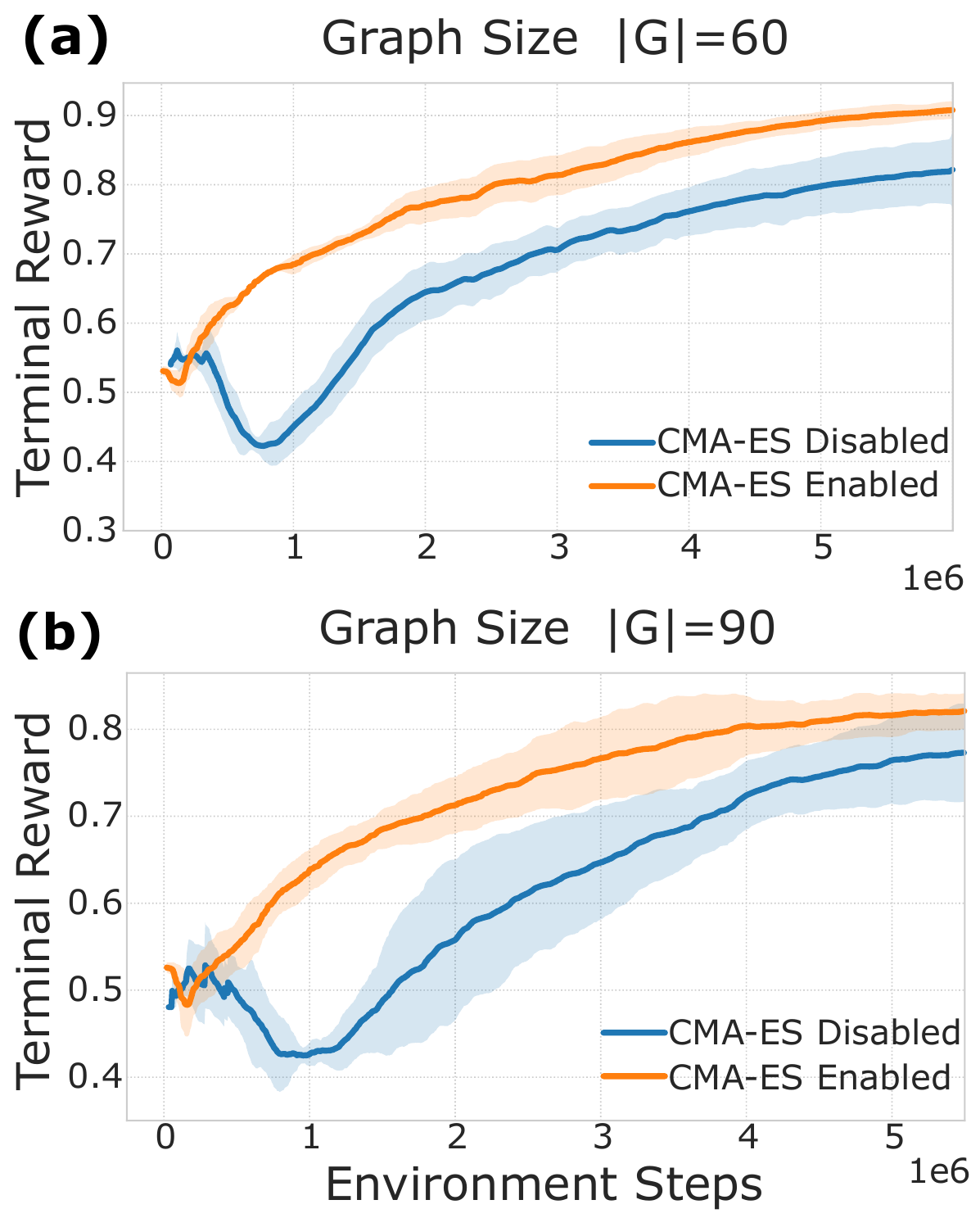}
        \captionof{figure}{Ablation study of action projection with CMA-ES enabled, vs. disabled for varying target graph sizes.}
        \label{fig:ablation-study}
    \end{minipage}
    \hfill
    \begin{minipage}[t]{0.53\linewidth}
        \centering
        \includegraphics[width=1\linewidth]{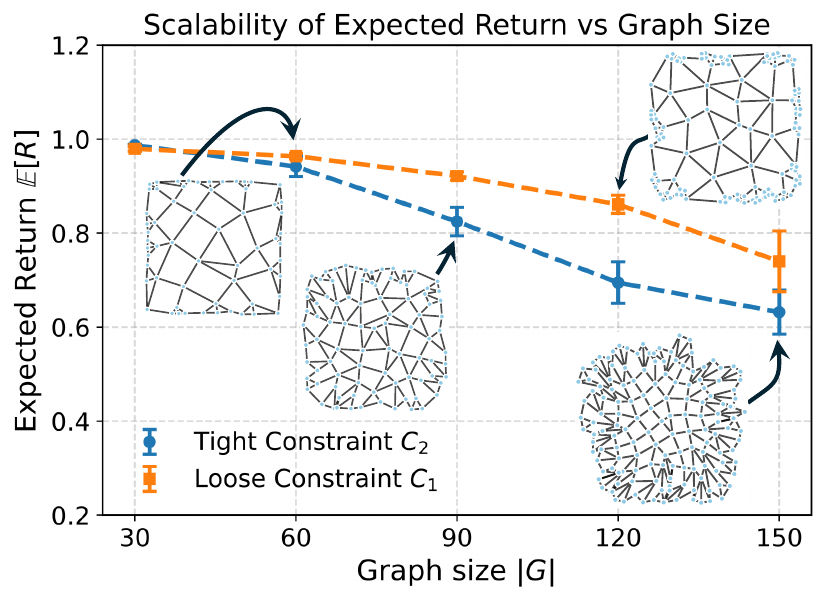}
        \captionof{figure}{Scalability study of our method for different graph sizes and constraint tightness.}
        \label{fig:scalability-study}
    \end{minipage}

    \vskip -0.2in
\end{figure}

We further evaluate action projection and scalability on a single squareness objective $r_\mathrm{square}$ over varying graph sizes. Figure~\ref{fig:ablation-study} shows that CMA-ES-supervised action projection improves terminal reward by steering the policy away from infeasible continuous actions, allowing the agent to focus on reward optimization. This also reduces constraint violations by a factor of 15 compared to the projection-free variant; detailed statistics are provided in Appendix~\ref{app:action_projection_results}. Figure~\ref{fig:scalability-study} evaluates the reward of graph sizes up to 150 nodes under two constraint regimes. As expected, tighter constraints and larger graph sizes reduce achieved reward. However, the method still generates feasible square-like graphs at graph sizes substantially larger than common graph diffusion models operating on planar graph datasets, which typically use at most 64 nodes~\citep{martinkus_spectre_2022, vignac_digress_2023}.

\section{Limitations \& Future Work}
While our experiments generate graphs with up to 150 nodes (Figure~\ref{fig:scalability-study}), scalability remains limited by the growing combinatorial grammar design space for longer decision horizons. This limitation should be viewed relative to deep generative baselines, which often lose validity or efficiency on larger graphs under strict structural constraints such as planarity, as shown in Appendix~\ref{app:extended_results}. Tackling this scalability ceiling is left for future work, where a principled approach could be the adoption of hierarchical reinforcement learning \cite{Sutton99} to reduce the effective decision horizon. A higher-level policy would select actions that define larger structural changes, while a lower-level policy would apply the existing grammar to maximize the intrinsic reward obtained by the high-level policy. Such an extension could also facilitate applications to constrained, goal-directed mesh generation, where substantially larger graphs are typically required. A second limitation is the structural prior imposed by the grammar: the method can only generate graphs expressible by the chosen rule set. We therefore do not claim completeness with respect to all planar embedded graphs, and results should be interpreted within the assumptions encoded by the grammar. At the same time, the rule set is also a strength in domain-specific settings, where hand-crafted rules encode system validity by design \cite{parish_procedural_nodate}. However, deriving grammars with improved formal coverage of planar graph embeddings is an important direction for future work.

\section{Conclusions}
We introduced a grammar-based reinforcement learning method and benchmark suite that, to our knowledge, is the first to combine goal-directed generation of embedded planar graphs with hard structural and geometric constraint enforcement. By formulating graph rewriting as a constrained Markov decision process, the method constructs topology and geometry jointly while enforcing feasibility through grammar-induced admissibility and learned action projection. Across four benchmarks, our method generates graphs with exact prescribed node counts, satisfies all constraints, and achieves the best overall multi-objective reward. Although a few baselines are competitive in the benchmarks, they do not generalize across other multi-objective tasks; by contrast, our method maintains strong rewards while satisfying the full constraint set in all benchmarks, indicating stronger generalization across heterogeneous benchmarks. On the other hand, diffusion-based generators struggle strongly to satisfy constraints while maximizing the reward. Beyond the metrics and constraints used here, the framework is designed to accommodate arbitrary user-specified objectives and constraints through the reward and constraint definitions.

\section*{Acknowledgments}
This research was supported by the ETH Research Grant 23-1 ETH-35. Research supported by the NVIDIA Academic Grant Program using a NVIDIA RTX PRO 6000 Blackwell Max-Q Workstation Edition GPU.

{
\small

\bibliographystyle{unsrtnat}
\bibliography{bib}
}

%%%%%%%%%%%%%%%%%%%%%%%%%%%%%%%%%%%%%%%%%%%%%%%%%%%%%%%%%%%%%%%%%%%%%%%%%%%%%%%
% APPENDIX
%%%%%%%%%%%%%%%%%%%%%%%%%%%%%%%%%%%%%%%%%%%%%%%%%%%%%%%%%%%%%%%%%%%%%%%%%%%%%%%
\newpage
\appendix
\section{Grammar Rules}
\label{app:rules}

This appendix specifies the seven graph rewriting rules defined in the grammar
$\mathcal{G}$.
Each rule $p\in\mathcal{P}$ acts on a single boundary node
$\bar v_t^b$ determined by the match $g_t$ and is applied via the graph transition operator $\mathcal{T}$ to the graph state $G_t$.
A rule may introduce new embedded nodes and edges connecting those to $\bar v_t^b$,
parameterized geometrically by a rule-specific continuous vector
$\theta_t\in\Theta(p_t)$ that encodes normalized sector angles and edge lengths (cf. Appendix~\ref{app:rule_parametrization}).
All rules update both the directed edge set $E_t^d$ and the node set $V_t$. Prior to each rule application, the boundary edges $E_t^b$ in $G_t$ are removed as part of $\mathcal{T}$, and reintroduced afterwards according to the connecting procedure (cf. Appendix~\ref{app:connecting_procedure}). This connecting procedure identifies how the leaf nodes in $V_t^b$ in the current graph state are connected using boundary edges, such that the resulting graph is planar and geometrically closed. These boundary edges then populate the edge set $E_t^b$. The rule semantics and notations are introduced in Appendix~\ref{app:rule_semantics}. A schematic overview of all rules in $\mathcal{P}$ is shown in Figure~\ref{fig:rules} and Table~\ref{tab:grammar-rules}. 

\begin{figure}[h]
    \vskip 0.2in
    \begin{center}
    \centerline{\includegraphics[width=\columnwidth]{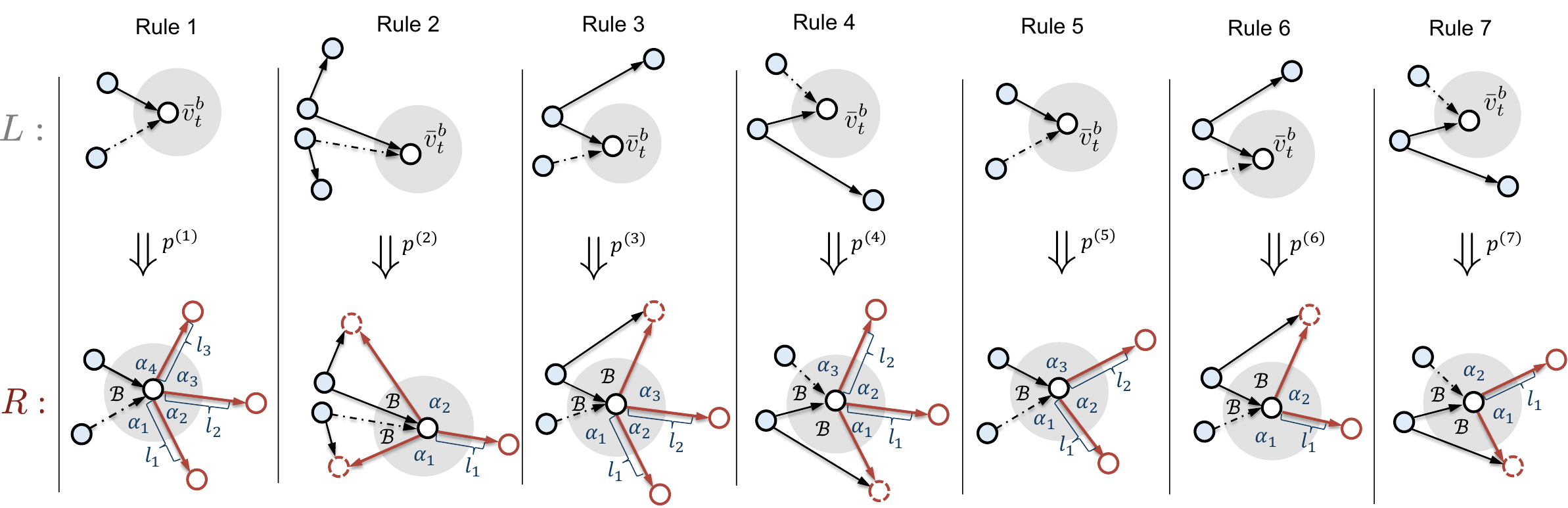}}
    \caption{Schematic illustration of the seven grammar rules in $\mathcal{P}$. The top row represent a graph state $G_t$ and the highlighted left-hand side $L$ (i.e. the match $\bar v_t^b$) of a rule $p^{(\cdot)}$ prior to a rule application. The bottom row shows the graph state $G_{t+1}$ after a rule has been applied, with the right-hand side $R$ shown as red outward directed edges, the red outlined newly added nodes, and merged nodes as dashed outlined nodes. Geometric parameters are indicated in the bottom row.}
    \label{fig:rules}
  \end{center}
\end{figure}

\subsection{Rule semantics and notation in Figure~\ref{fig:rules} and Table~\ref{tab:grammar-rules}.}
\label{app:rule_semantics}
Each rule $p^{(\cdot)} \in \mathcal{P}$ .
With reference to Section~\ref{sec:graph_grammar}, a rule $p^{(\cdot)}: L \to R$ rewrites a local neighborhood (left-hand side $L$) in the graph state $G_t$ around a boundary node $\bar v_t^b$ identified by the match $g_t$ with a new subgraph $R$. In the visualizations of all rules in Figure~\ref{fig:rules}, the top row depicts the graph state $G_t$ prior to a rule application with the highlighted subgraph $L$ (i.e. the match $\bar v_t^b$), which can be rewritten with a rule and the graph transition operator $\mathcal{T}$. Additionally, we depict optional incoming directed edges to $\bar v_t^b$ using dashed edges. This represents edges that may, or may not be present at the rule application site, implying that any number of existing incoming directed edges may be valid at a selected match $\bar v_t^b$, as long as the degree constraint $c_{\deg}$  (see Section~\ref{sec:contax}) is satisfied.

The bottom row in Figure~\ref{fig:rules} depicts the graph state $G_{t+1}$ where the right-hand side $R$ of the rules has been applied. The newly added edges are depicted as outward directed edges from $\bar v_t^b$ in red with new nodes with a red outline. Several rules allow to merge an edge to an existing node in the graph $G_t$ (i.e. rules 2,3,4,6,7). This merging mechanism is enabled by the connecting procedure (cf. Appendix~\ref{app:connecting_procedure}). Figure~\ref{fig:rules} highlights the merged nodes in $R$ as dashed-outlined nodes.

Each rule is associated with a parameter space $\theta_t\in\Theta(p_t)$ that defines the geometric parametrization of angles $\alpha_i$ (sector angles) and lengths $\ell_i$ (line
segments) in $R$. In the bottom row of Figure~\ref{fig:rules}, we depict the set of sector angles occupied by edges that are present at $\bar v_t^b$ prior to rule applications using the set $\mathcal{B}$. Hence, the angular parameter space of $\Theta(p_t)$ is restricted to the remaining angular budget $B = 2\pi - \sum_{k\in\mathcal{B}} \alpha_k$.

Boundary edges $E_t^b$ are not depicted in Figure~\ref{fig:rules}, as they are removed before any rule application by the transition operator $\mathcal{T}$ and reintroduced afterward by use of the connecting procedure (cf. Appendix~\ref{app:connecting_procedure}). Note that, when discarding the undirected boundary edge set, the described rule-set creates Directed Acyclic Graphs (DAG) by design, since rules are applied uniquely at leaf nodes in a directed graph. Additionally, merge-able nodes, identified by the connecting procedure are uniquely leaf nodes, too, therefore, the application of the rewriting rules cannot introduce directed cycles and the grammar $\mathcal{G}$ will always produce a DAG.

\begin{table}[h!]
\caption{Summary of grammar rules $\mathcal{P}$. “Merge” refers to connections identified by
the connecting procedure (cf. Figure~\ref{fig:marching}) and "Sprout" refers to newly added nodes. The table additionally shows the dimensionality of the geometric parametrization.}
\label{tab:grammar-rules}
\centering
\vspace{0.35em}
\begin{tabular}{clcccc}
\toprule
Rule set $\mathcal{P}$ & Description & New nodes & Merges & $(\#\alpha,\#\ell)$ & $\dim\Theta$ \\
\midrule
$p^{(1)}$ & Triple node sprout                     & 3 & 0 & $(4,3)$ & 7 \\
$p^{(2)}$ & Single sprout + two merges              & 1 & 2 & $(2,1)$ & 3 \\
$p^{(3)}$ & Merge left + double sprout              & 2 & 1 & $(3,2)$ & 5 \\
$p^{(4)}$ & Merge right + double sprout             & 2 & 1 & $(3,2)$ & 5 \\
$p^{(5)}$ & Double node sprout                      & 2 & 0 & $(3,2)$ & 5 \\
$p^{(6)}$ & Merge left + single sprout              & 1 & 1 & $(2,1)$ & 3 \\
$p^{(7)}$ & Merge right + single sprout             & 1 & 1 & $(2,1)$ & 3 \\
\bottomrule
\end{tabular}

\end{table}

\begin{figure}[t]
    \begin{center}
    \centerline{\includegraphics[width=0.5\columnwidth]{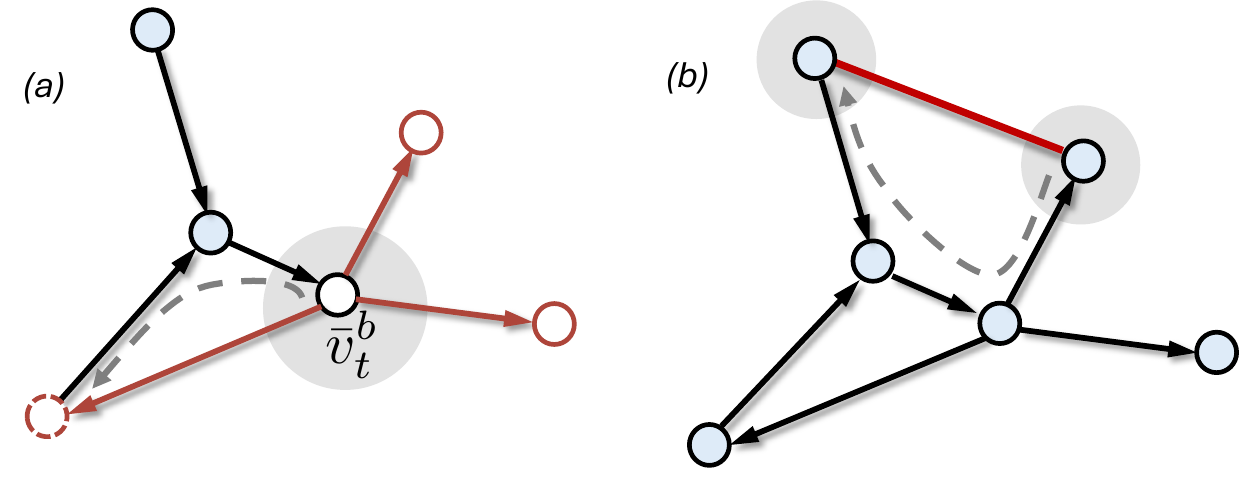}}
    \caption{Visualization of the connecting procedure. (a) The connecting procedure identifies a merge-able node, starting from $\bar v_t^b$. The merge-able node is depicted as the dashed-outlined node. (b) The connecting procedure identifies a boundary edge, depicted by the undirected edge between the highlighted nodes.}
    \label{fig:marching}
  \end{center}
\end{figure}

\subsection{Connecting Procedure cf. Figure~\ref{fig:marching}}
\label{app:connecting_procedure}

The \emph{connecting procedure} is a deterministic geometric routine that serves two purposes: (i) it identifies merge-able nodes that can be connected to from a match $\bar v_t^b$ by rules with merging semantics, while ensuring topological planarity (cf. Figure~\ref{fig:marching}(a)), (ii) it  reconstructs the graph's boundary edges after each rule application so that the updated graph remains planar and geometrically closed (cf. Figure~\ref{fig:marching}(b)).

The procedure operates exclusively on the current embedded graph state and depends only on the node positions and the directed edge configuration $E_t^d$. In case (i), the procedure is initialized at the match $\bar v_t^b$ and geometrically traverses the connected edges in $E_t^d$, starting from $\bar v_t^b$, in either clock-wise or counter-clock-wise direction, depending on the selected rule. The traversal terminates upon finding a boundary node, which is then labeled as the node to merge to. This is indicated in the figure using the dashed-curved arrow (traversal process) and the dashed-outlined node (merge-able node) in Figure~\ref{fig:marching}(a). In case (ii), the procedure is initialized at any boundary node and follows the same traversal principle until another boundary node is found. Then, a boundary edge connects both boundary nodes, depicted as the undirected edge between the highlighted nodes in Figure~\ref{fig:marching}(b). Applying this process to all boundary nodes guarantees that the resulting boundary edge set $E_t^b$ forms a non-self-intersecting polygonal chain that encloses the graph.

\subsection{Rule Parametrization}
\label{app:rule_parametrization}
In the implementation, $\mathcal{T}(G_t;g_t,p_t,\theta_t)$ maps the normalized parameters
$\theta_t\in(-1,1)^{\dim(\Theta(p_t))}$ to Cartesian coordinates.
For each rule $p_t$, the parameter space $\Theta(p_t)$ is decomposed into angular
parameters $\Theta_\alpha(p_t)$ and length parameters $\Theta_\ell(p_t)$, which
are denormalized independently using the procedures described below.

\paragraph{Sector angles.}
Rewrite rules introduce a finite set of new sector angles around $\bar v_t^b$.
Let $\mathcal{B}$ denote the set of angles already occupied by incoming directed
edges at $\bar v_t^b$ (cf. Figure~\ref{fig:rules}), and let
$B = 2\pi - \sum_{k\in\mathcal{B}}\alpha_k$ be the remaining angular budget.
For a rule that requires explicit angles $n$ plus one closing angle, let $m=n+1$ and
let $\theta_\alpha\in(0,1)^m$ denote the normalized angular parameters obtained by
rescaling the policy outputs from $(-1,1)$.
A minimum angle $\alpha_{\min}>0$ is enforced by reserving $\alpha_{\min}$ per
angle, yielding free budget
$B_{\mathrm{free}}=\max(B-m\alpha_{\min},0)$.
Denormalized angles are computed as
\begin{equation}
u_i=\theta_{\alpha_i}^2,\qquad
\tilde u_i=\frac{u_i}{\sum_j u_j+\delta}\,B_{\mathrm{free}},\qquad
\alpha_i=\alpha_{\min}+\tilde u_i,
\end{equation}
with $\delta>0$ a small constant.
This guarantees $\alpha_i\ge\alpha_{\min}$ and $\sum_i\alpha_i\le 2\pi$, with the
final angle acting as a closing angle.

\paragraph{Edge lengths.}
Each rule specifies a fixed number $\hat L$ of newly introduced edges and thus
$\hat L$ corresponding length parameters
$\theta_\ell\in(0,1)^{\hat L}$ obtained by rescaling the policy outputs from $(-1,1)$.
Physical edge lengths are given by
\begin{equation}
\ell=\ell_{\min}+\theta_\ell(\ell_{\max}-\ell_{\min}),
\end{equation}
where $\ell_{\min}$ and $\ell_{\max}$ are global bounds from the constraint set
$\mathcal{C}$.

\paragraph{Geometric realization.}
Let $\hat r$ denote the incoming reference direction at $\bar v_t^b$, conventionally taken as the first incident directed edge in counter-clockwise direction that sits within the polygon formed by the boundary edges $E_t^b$, and $R(\alpha)$ a planar
rotation by angle $\alpha$ with respect to the last incident edge in counter-clockwise direction within the polygon $E_t^b$.
All newly introduced vertices are placed in the Euclidean plane as
\begin{equation}
y_{\mathrm{new}}
=
y_{\bar v_t^b}+\ell\,\frac{R(\alpha)\hat r}{\|R(\alpha)\hat r\|_2}.
\end{equation}

\subsection{Implementation}
The complete grammar and graph rewriting logic is implemented in \textsc{JAX} \citep{jax2018github}, enabling compiled and efficient implementation of the transition operator. Low-level geometric operations, in particular planarity checks based on pairwise edge-intersection tests, are formulated in a vectorized manner and evaluated in parallel. The grammar execution runs on the CPU, as the control flow  of the grammar is predominantly sequential with sparse data structures that benefit more from high multi-core CPU throughput than from GPU parallelism.

\section{Actor--Critic Architecture}
\label{app:actor_critic}
In this appendix, we describe the complete actor--critic architecture in detail. The whole architecture is parametrized with neural networks, denoted with $\phi$.

\subsection{Graph Feature Extractor}
\label{app:feature_extractor}
First we describe the geometry-aware graph feature extractor used by the
actor--critic networks.
The extractor maps an embedded graph state $G_t$ to node-level and graph-level
representations suitable for decision-making, while respecting the geometric
structure of the graph.

\paragraph{Input representation.}
The extractor operates on a batched graph representation.
Each node $v \in V_t$ is associated with a feature vector
\begin{equation}
    x_v = [\, b_v,\ d_v,\ h_v,\ y_v \,],
\end{equation}
where $b_v\in\{0,1\}$ is a boundary indicator, $d_v\in\{0,\dots,D_{\max}\}$ is the
(discrete) degree, and $h_v\in\mathbb{R}^{2K}$ is a precomputed sector-angle
embedding using $K=4$ Fourier harmonics (cf. Figure~\ref{sec:feature_extraction}).
Each node additionally carries planar coordinates $y_v\in\mathbb{R}^2$.
Edges are treated as undirected for the purpose of message passing.

\subsubsection{Node-level Embeddings}
\label{app:node_level_embeddings}

\paragraph{Discrete and continuous feature lifting.}
Discrete degree values are embedded via a learned embedding (lookup table)
$e(d_v)\in\mathbb{R}^{\mathcal{E}_d}$.
The discrete block $[\,b_v,\ e(d_v)\,]$ is concatenated with the continuous
sector-angle features $h_v$ and mapped to the hidden dimension $Z$ using a
two-layer perceptron $\phi_{\mathrm{lift}}$,
\begin{equation}
    \tilde h_v^{(1)} = \phi_{\mathrm{lift}}\!\left(h_v \,\|\, b_v \,\|\, e(d_v)\right) \in \mathbb{R}^Z,
\end{equation}
while remaining E(2)-invariant. Initial coordinates are taken as
$y_v^{(1)} = y_v$.

\paragraph{EGNN backbone.}
The lifted features are processed by a stack of three E(2)-equivariant graph
neural network (EGNN) layers~\citep{satorras_en_2022}.
At each layer $l$, E(2)-invariant node features $\tilde h_v^{(l)}$ and
E(2)-equivariant coordinates $y_v^{(l)}$ are jointly updated using relative
positions encoded via a fixed Fourier encoding with four frequencies.
Each layer applies residual connections followed by layer
normalization on the invariant features, yielding updated node features
$\tilde h_v^{(l+1)}$.
Coordinate updates $y_v^{(l+1)}$ are not affected by residual connections or
normalization.
The backbone produces a sequence of node-feature tensors
$\{\tilde H^{(1)}, \tilde H^{(2)}, \tilde H^{(3)}\}$, where
$\tilde H^{(l)}\in\mathbb{R}^{|V_t|\times Z}$ and $|V_t|$ vary with the current
graph size.
Coordinate updates are equivariant under global planar rigid motions, while node
and graph features are invariant to translations and rotations.
All global readouts are permutation-invariant with respect to node ordering.

\paragraph{Graph-size-aware jumping knowledge.}
To adapt representations across varying graph sizes, the extractor employs a
graph-size-aware jumping-knowledge mechanism \citep{xu_representation_2018}.
For each layer $l$, mean-pooled features
$g_\mathrm{pool}^{(l)} = \frac{1}{|V_t|}\sum_v \tilde h_v^{(l)}$
are concatenated with a log-scaled graph-size signal $\log(1+|V_t|)$ that automatically bounds the graph size for a well-behaved MLP input.
A two-layer MLP produces normalized layer weights $w^{(l)}_{\mathrm{gate}}$, which are used
to fuse per-layer node feature tensors into the final node embedding tensor 
\begin{equation}
\label{eq:node-level-features}
    \hat H = \sum_{l=1}^{3} w^{(l)}_{\mathrm{gate}}\, \tilde H^{(l)} \in \mathbb{R}^{|V_t| \times Z},
\end{equation}
where each row $\hat H_v$ in $\hat H$ corresponds to a node embedding at node $v \in V_t$.

\subsubsection{Graph-level Embeddings}
Graph-level embeddings $\hat g \in \mathbb{R}^{Z}$ are obtained by summing attentional aggregation
\citep{li_graph_2019}, implemented via a two-layer gating MLP, and mean pooling
over the set of node embeddings $\{\hat H_v\}_{v=1}^{|V_t|}$.

\subsubsection{Feature Aggregation}
As described in Section~\ref{sec:feature_extraction}, for downstream policy network inputs we
concatenate each node embedding $\hat H_v$ with the graph-level embedding $\hat g$ 
\begin{equation}
    H_v = [\hat H_v \| \hat g] \in \mathbb{R}^{2Z}
\end{equation}
to form a final set of fused node-level representations
$\{H_v\}_{v=1}^{|V_t|}$.
Including graph-level information enables the agent to make globally informed
decisions using only node-level features $H_v$ while preserving the Markov property.

For the critic, only the graph-level representation $\hat g$ is used as input,
and node-level representations are discarded.

\subsection{Policy Networks}
\label{app:policy_networks}
We describe the neural networks used to model the policy distributions
$\pi_\phi(g_t\mid G_t)$ (matching),
$\pi_\phi(p_t\mid g_t, G_t)$ (rule selection),
and $q_\phi(\xi_t\mid g_t, p_t, G_t)$ (parametrization), as introduced in Section~\ref{sec:agent_design}.

\subsubsection{Matching}
The matching distribution $\pi_\phi(g_t\mid G_t)$ is approximated by a two-layer
MLP with domain $\{\mathbb{R}^{2Z}\}^{|V_t|} \to \mathbb{R}$, taking as input the
node representations $\{H_v\}_{v=1}^{|V_t|}$ and producing one logit per node.
By definition of admissible matches (cf. Figure~\ref{sec:graph_grammar}), only boundary
nodes are valid; all non-boundary nodes are therefore masked by assigning
$-\infty$ logits.
The resulting logits define the categorical distribution
$\pi_\phi(g_t\mid G_t)$, from which a boundary node $\bar v_t^b$ is sampled.

\subsubsection{Rule Selection}
The rule-selection distribution $\pi_\phi(p_t\mid g_t, G_t)$ is modeled by a
two-layer MLP $\mathbb{R}^{2Z} \to \mathbb{R}^{P_{\max}}$ with input
$H_{\bar v_t^b}$. Here, $P_{\max} = 7$, corresponding to the number of rules implemented (cf. Table~\ref{tab:grammar-rules}).
Conditioning on $(g_t,G_t)$ is thus realized by evaluating the node embedding at
the sampled match location $\bar v_t^b$ with $H_{\bar v_t^b}$.
The network outputs one logit per rule in $\mathcal{P}$, defining the categorical
distribution $\pi_\phi(p_t\mid g_t, G_t)$ from which a rule $\bar p_t$ is sampled.

\subsubsection{Parametrization}
\label{app:parametrization_policy}
The continuous parametrization
$q_\phi(\xi_t\mid g_t, p_t, G_t)=\mathcal{N}(\mu_\phi,\sigma_\phi^2)$
is modeled as an isotropic Gaussian.
The mean $\mu_\phi$ is conditioned on the sampled rule via a learned rule
embedding $e(p_t)\in\mathbb{R}^{\mathcal{E}_p}$ and predicted by a two-layer MLP with domain
$\mathbb{R}^{2Z}\times\mathbb{R}^{\mathcal{E}_p}\to\mathbb{R}^{D_{\max}}$, taking inputs
$H_{\bar v_t^b}$ and $e(p_t)$. Here, $D_{\max} = \max_{p\in\mathcal{P}} \dim \Theta(p) = 7$, corresponding to the maximum number of continuous parameters for all rules. For any rule that requires fewer parameters, we apply masking to filter unused dimensions.
The log-variance $\log\sigma_\phi^2$ is a global trainable vector of dimension
seven with no state-dependent inputs.
This Gaussian represents the nominal parametrization $\xi_t$ with infinite
support prior to the action projection.

\subsection{Critic Network}
The critic predicts a scalar state value from the graph-level embedding $\hat g_t \in \mathbb{R}^Z$
and discards node-level representations. Note that the critics feature extractor does \emph{not} share the feature-extractor parameters with the policies feature extractor. Each
uses an architecturally identical but independently trained extractor, to disentangle learning signals for actor and critic.
To provide an explicit graph-size signal, a learned embedding
$e(\log(1+|V_t|))\in\mathbb{R}^{Z}$ is added to form
$\tilde g_t=\hat g_t+e(\log(1+|V_t|))$.
The state value $V(G_t)$ is then predicted from $\tilde g_t\in\mathbb{R}^Z$ using
a two-layer MLP.

\subsection{Implementation, Network Sizes, and Training}
The complete actor--critic architecture is implemented using
\textsc{PyTorch} and \textsc{PyTorch-Geometric}~\citep{fey_fast_2019}, while the
environment is implemented in \textsc{JAX} \citep{jax2018github}.
The PPO implementation is inspired by, but independently re-implemented from
\textsc{StableBaselines3}~\citep{stable-baselines3} to support graph
data structures.

\paragraph{Network Sizes.}
All mentioned actor--critic MLPs use two-layers (\texttt{Linear - SiLU - Linear}) (including MLPs within the EGNN layers) with \texttt{SiLU} activations.
The hidden dimension $Z=32$ is used throughout, except for the critic MLP, which uses
dimension $64$.
The degree and rule embeddings are set to $E_d=E_p=20$.
In total, the architecture contains $66{,}972$ trainable parameters,
illustrating the expressivity of the model with only a few tens of thousands of parameters.

\paragraph{PPO.}
PPO optimization uses the \textsc{Adam} optimizer~\citep{kingma_adam_2017} with a
cosine-annealed learning-rate schedule.
Training is performed on an NVIDIA RTX PRO 6000 Blackwell Max-Q GPU, while
the environment runs on a 16-core AMD CPU with multiprocessing.
Training time ranges from 10 minutes to 2 hours depending on the task complexity.
Hyperparameters are listed in Table~\ref{tab:ppo_hyper}.
All policy networks, including feature extractors, are orthogonally initialized
with gain $\sqrt{2}$; final linear layers of matching, rule-selection, and
parametrization networks use gain $0.01$ and zero bias to encourage early
exploration.
The log-variance vector $\log\sigma_\phi^2$ is initialized to zero
($\sigma=1$), and critic networks are initialized with gain $1.0$.

\begin{table}[h!]
  \caption{PPO hyper-parameters chosen consistently for all experiments.}
  \label{tab:ppo_hyper}
\begin{center}

\begin{tabular}{l c}
\hline
Parameter & Value \\
\hline
Discount factor $\gamma$ & 0.99 \\
GAE parameter $\lambda$ & 0.95 \\
Clip range $\epsilon_{\mathrm{PPO}}$ & 0.2 \\
Entropy coefficient $c_{\text{ent}}$ & 0.01 \\
PPO epochs per update & 5 \\
Mini-batch size & 675 \\
Number of mini-batches & 4 \\
Number of parallel environments & 30 \\
Number of steps per rollout & 90 \\
Time limit for episode & 400 \\
Learning rate & $4\times10^{-4}$ (decay to $10^{-5}$) \\
Target KL-divergence & 0.02 \\
\hline
\end{tabular}
\end{center}
\end{table}
\section{Action Projection}
\label{app:action_projector}

This appendix describes the learnable action projection module $\eta_\psi$ used
to map nominal continuous action parameters to a state-dependent feasible set, as
introduced in Section~\ref{sec:agent_design} and Section~\ref{sec:safe_action_projection_learning}.
We detail (i) the action projection architecture, (ii) the supervised training procedure,
and (iii) the generation of supervision signals for action projection training. We additionally show that the policy space is preserved up to a reparameterization using the action projection in Appendix~\ref{sec:policy_space_preservation}

\subsection{Action Projection Architecture}
\label{app:projector_architecture}

At time step $t$, the actor produces a nominal hybrid action
$a_t=(g_t, p_t, \xi_t)$, where $g_t$ denotes the selected boundary node (match),
$ p_t\in\mathcal{P}$ the selected grammar rule, and
$ \xi_t$ a nominal continuous parameter vector.
The action projection is then conditioned on the local node-level embedding at the selected match and on the discrete rule choice using a learned embedding $e(p_t)\in\mathbb{R}^{\mathcal{E}_p}$. The sampled node-level embedding $\hat H_{\bar v_t^b}\in\mathbb{R}^{Z}$ is extracted using the equivalent feature extractor (cf. Appendix~\ref{app:feature_extractor} and Section~\ref{sec:feature_extraction}) as for the actor--critic networks, but trained separately in terms of trainable parameters $\psi$.

\subsubsection{Squashed Affine Map}
Conditioned on the concatenation of a sampled node embedding $\hat H_{\bar v_t^b}$ and $e(\bar p_t)$ the projection uses an element-wise affine transformation followed by a
$\tanh$ squashing nonlinearity, explicitly formulated as
\begin{equation}
\label{eq:app_projector_map}
\theta_t
=
\eta_\psi(\xi_t;G_t,g_t,p_t)
=
\tanh\!\Big(
s_\psi(H_{\bar v_t^b}\|e(\bar p_t))\odot \xi_t + t_\psi(H_{\bar v_t^b}\|e(\bar p_t))
\Big)
\end{equation}
where $s_\psi(\cdot)>0$ and $t_\psi(\cdot)$ denote learned scale and shift
functions, respectively, and $\odot$ denotes element-wise multiplication.

The scale and shift functions $s_\psi(\cdot)>0$ and $t_\psi(\cdot)$ are modeled as a 2-layer MLP $\mathbb{R}^{Z+\mathcal{E}_p} \to \mathbb{R}^{D_{\max}}$ with input $[\,H_{\bar v_t^b} \,\|\, e(p_t)\,]$ each. Additionally, the output of the scale MLP $s_\psi(\cdot)$ is $\log s_\psi$ bounded via
$\log s_\psi=\tanh(\cdot)\,\kappa$ with $\kappa=2$, yielding element-wise scale
values $s_\psi\in[\exp(-2),\exp(2)]$.

The final linear layers of both MLPs are initialized to zero, resulting in an initial
near-identity mapping (unit scale and zero shift) up to the outer $\tanh$. This enforces unbiased initial mapping before training the action projection and reflects a standard squashed Gaussian projection. Hidden layer dimensions are set to 32.

\subsubsection{Jacobian determinant}
Since Equation~\ref{eq:app_projector_map} is element-wise, the Jacobian is diagonal and
the log absolute determinant is available in closed form,
\begin{equation}
\log\bigl|\det J_{\eta_\psi}\bigr|
=
\sum_{i=1}^{D_{\max}}
\mathrm{mask}_i(p_t)\,\Big(
\log s_{\psi,i}
+
\log\bigl(1-\theta_{t,i}^2\bigr)
\Big),
\end{equation}
with numerical stabilization obtained by clamping $\theta_{t,i}$ away from
$\pm1$ with a small $\varepsilon$. $\mathrm{mask}_i(p_t)$ represents masking for unused rule-dependent dimensions.
During PPO optimization, this term is used to apply the standard
change-of-variables correction (cf. Section~\ref{sec:agent_design})
\begin{equation}
    \log\pi_\phi(\theta_t\mid\cdot)
=
\log q_\phi(\xi_t\mid\cdot)
-
\log\bigl|\det J_{\eta_\psi}\bigr|.
\end{equation}

\subsubsection{Policy Space Preservation}
\label{sec:policy_space_preservation}
We show that the policy space $\Theta$ is preserved up to a reparameterization using the action projection $\eta_\psi$.

Let $\Xi = \mathbb{R}^d$ and $\Theta = (-1,1)^d$.
The map $\eta_\psi : \Xi \to \Theta$, as specified in Equation~\ref{eq:app_projector_map}, is a $C^1$ diffeomorphism.

Define the push-forward operator
\begin{equation}
    \eta_{\psi\#} : \hat{\mathcal{P}}(\Xi) \to \hat{\mathcal{P}}(\Theta),
\qquad
\eta_{\psi\#}  \mu := \mu \circ \eta^{-1}_\psi,
\end{equation}

where $\hat{\mathcal{P}}(\cdot)$ denotes the set of probability measures.
Then, $\eta_{\psi\#} $ is a bijection with inverse $(\eta^{-1}_\psi)_{\#}$. Consequently, the policy classes
\begin{equation}
    \{\; q_\phi \in \hat{\mathcal{P}}(\Xi) \;\}
\quad\text{and}\quad
\{\; \pi_\phi = \eta_{\psi\#} q_\phi \in \hat{\mathcal{P}}(\Theta) \;\}
\end{equation}
are isomorphic via the push-forward induced by the diffeomorphism $\eta_{\psi}$, and the policy space is preserved up to reparameterization.

\subsection{Supervised Training Procedure}
\label{app:projector_training}

The action projection is trained using a short FIFO replay buffer that stores recent
graph transitions collected during PPO rollouts. Since the action projection is off-policy in parameters $\psi$, we store the last ten rollouts in the replay buffer and train it synchronously alongside PPO.
Each entry in the buffer consists of graph states together with nominal continuous parameters
$\xi_t$ and the corresponding environment-projected targets $\theta_t^\star$ using Covariance Matrix Adapted Evolutionary Strategy (CMA-ES)
(see Appendix~\ref{app:cmaes_aggregation}).

Training minimizes a mean-squared error loss over the continuous parameters,
\begin{equation}
\label{eq:loss_action_projection}
\mathcal{L}_{\mathrm{proj}}
=
\mathbb{E}\Big[
w_t\;
\bigl\|
\mathrm{mask}(p_t)\odot(\eta_\psi(\xi_t)-\theta_t^\star)
\bigr\|_2^2
\Big],
\end{equation}
where the sample weight $w_t$ is set to $1$ if the environment modified any
continuous parameter, and to $0.5$ otherwise.
This biases learning toward the cases where geometric correction is required.

The projection is optimized using the same optimizer and learning-rate schedule
as PPO. The projection is then trained for a fixed number of 7 supervised epochs per PPO iteration.
After a prescribed number of 200 PPO iterations, the projection training is terminated,
and the projection remains fixed during subsequent policy optimization.

\subsubsection{Environment-Side Supervision via CMA-ES}
\label{app:cmaes_aggregation}

For each step, the actor produces nominal samples $\xi_t$ and projects them using the action projection $\eta_\psi$ to $\theta_t$. The environment then assesses the feasibility of the proposed continuous parameters $\theta_t$ using geometric $\mathcal{C}_G$ and topological $\mathcal{C}_T$ constraints checks (cf. Section~\ref{sec:contax}).
If an action is topologically admissible but geometrically infeasible (e.g., due
to edge intersections or boundary length violations), the CMA-ES
procedure is invoked (cf. Section~\ref{sec:safe_action_projection_learning}) to further correct the proposed $\theta_t$. If CMA-ES finds an admissible solution $\theta_t^\star$, the action projection will be updated and trained using the corrected solution in the next training iteration.

CMA-ES is initialized at the proposed parameters and minimizes an objective (cf. Equation~\ref{eq:cmaes_penalty_objective}) that encourages proximity to the original parameters while satisfying $\mathcal{C}_G$.
If a feasible $\theta_t^\star$ is found, it is applied to advance the environment
state, otherwise we fall back to the rejection-sampling scheme while advancing the time step $t \to t+1$ without changing the environment state. Advancing the time step informs the actor that the action was infeasible implicitly, due to a reduced terminal reward according to the discount factor $\gamma < 1$.

The optimization problem stated in Equation~\ref{eq:cmaes} is implemented using an explicit penalty objective as
\begin{equation}
\label{eq:cmaes_penalty_objective}
J(\theta_t')
=
\lambda_{\mathrm{geom}}
\big(
c_{\mathrm{pl}}(\theta_t')
+
c_{{\ell}}(\theta_t')
\big)
+
\|\theta_t' - \bar\theta_t\|_2,
\end{equation}
with penalty weight $\lambda_{\mathrm{geom}}=10$, $c_{\mathrm{pl}}$ representing the planarity constraint in the form of edge-intersections and $c_{\ell}$ the boundary edge length constraint. 
Other constraints in $\mathcal{C}_G$ do not appear explicitly, as they are either
enforced by the rule parametrization (sector angles, internal edge lengths), or correspond
to violations introduced by merging rules, which are classified by topological constraints $\mathcal{C}_T$ and cannot be resolved by continuous reparameterization.
Since CMA-ES operates in an unconstrained space, optimization is performed over
$z\in\mathbb{R}^{\dim\Theta(p_t)}$ using the sigmoid reparameterization
$\theta_t'=\sigma(z) \in (0,1)^{\dim\Theta(p_t)}$ to operate in the rule parameterization space (cf. Appendix~\ref{app:rule_parametrization}), yielding
\begin{equation}
\min_{z\in\mathbb{R}^{\dim\Theta(p_t)}} \; J\!\bigl(\sigma(z)\bigr).
\end{equation}

CMA-ES is implemented using \textsc{Evosax} \citep{lange_evosax_2022} and run using a population size 100 with 4 iterations in total, an initial search standard deviation of 1.

\subsection{CMA-ES \& Action Projection Results}
\label{app:action_projection_results}

\begin{figure}[t]
  \begin{center}
    \centerline{\includegraphics[width=\columnwidth]{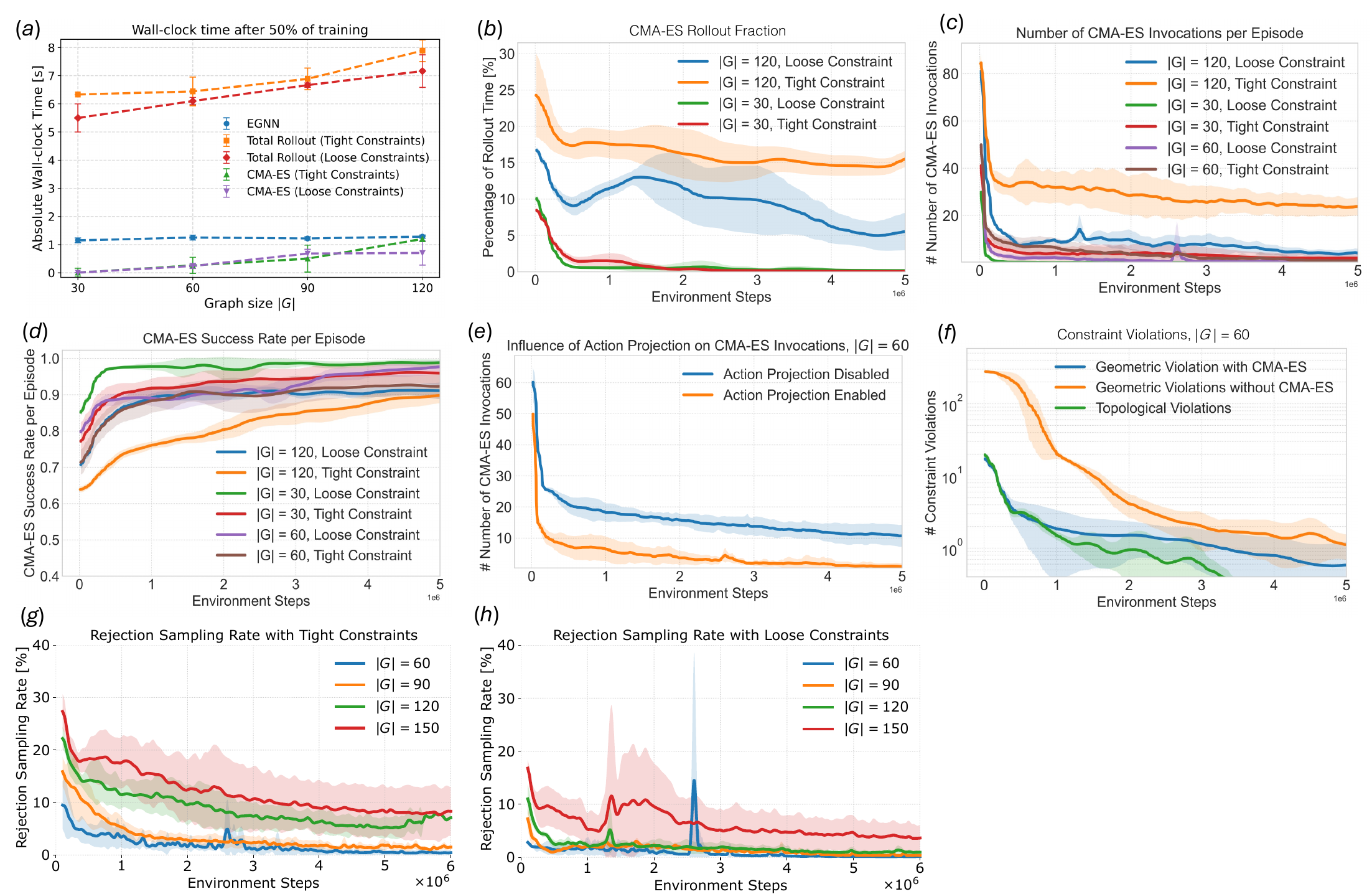}}
    \caption{Impact of CMA-ES optimization with action projection on training performance. Constraint tightness refers to the same constraint set as described in Section~\ref{sec:Experiments}. All error-bars refer to 1-sigma standard deviations accross 5 seeds.
    (a) Wall-clock time per rollout at 50\% of training for different graph sizes,
    with contributions from feature extraction (EGNN) and CMA-ES under loose and tight constraints.
    (b) Percentage of rollout time spent in CMA-ES during training.
    (c) Number of CMA-ES invocations per episode for different graph sizes and constraint tightness.
    (d) Success rate of CMA-ES in resolving geometric constraints.
    (e) CMA-ES invocations with learned vs. static projection ($s_\psi=1$, $t_\psi=0$).
    (f) Number of topological ($C_T$) and geometric ($C_G$) constraint violations
    during PPO (log scale).
    (g, h) Rejection sampling rate during training using tight and loose constraints.}

    \label{fig:action_proj_summary}
  \end{center}
\end{figure}

We further analyze the effect of CMA-ES and action projection on rollout wall-clock time in Figure~\ref{fig:action_proj_summary}(a,b), and report the number of CMA-ES invocations during training across graph sizes and constraint regimes in Figure~\ref{fig:action_proj_summary}(c). Figures~\ref{fig:action_proj_summary}(a) shows that, after 50\% of training, the computational overhead introduced by CMA-ES is negligible relative to the remaining rollout computations, while (b) shows the fractional proportion of CMA-ES during the whole training. This low overhead is due in part to the action projection mechanism, which progressively approximates the safe set during training, and renders CMA-ES only as a fallback mechanism, and to the efficient \textsc{Evosax} implementation. Consistently, Figure~\ref{fig:action_proj_summary}(c) shows that CMA-ES invocations rapidly decrease over training and become rare, partly due to the learned action projection. This effect is further supported by Figure~\ref{fig:action_proj_summary}(e), where we compare geometric constraint violations with action projection enabled against a static projection variant obtained by setting $s_\psi=1$ and $t_\psi=0$. Figure~\ref{fig:action_proj_summary}(d) reports the CMA-ES success rate and shows that, even for large graphs under tight constraints, the optimizer finds feasible parameters $\theta_t^\star$ satisfying $\mathcal{C}_G=0$ at a high rate. Further, as discussed in Section~\ref{sec:safe_action_projection_learning}, topological violations are substantially less frequent than geometric violations, as shown on the logarithmic scale in Figure~\ref{fig:action_proj_summary}(f). Finally, Figure~\ref{fig:action_proj_summary}(g) and (h) shows the overall rejection sampling rate, which comprises both topological and geometric constraint violations after action projection, which then lead to rejection sampling. For smaller graph sizes this rate quickly converges to zero, while larger graphs with tighter constraints more often reside to rejection sampling. Nevertheless, this rate still remains low.

\newpage
\section{Graph Metrics}
\label{app:metrics}

We describe the graph metrics used in the experiments in Section~\ref{sec:Experiments}. Metrics are categorized as topological (adjacency-based), geometry-based (drawing-dependent) or spectral (Laplacian-based). All metrics are summarized in Table~\ref{tab:metrics_summary}.

\subsection{Topological and Spectral Metrics}
The topological and spectral metrics described here are enumerated and discussed with more detail in \citep{drobyshevskiy_random_2020, kunegis_guided_2023}

For all metrics, we represent the union of internal and boundary edges $E_t^d \cup E_t^b$ as an undirected graph with adjacency matrix
\( A \in \{0,1\}^{n \times n} \).

% \textbf{Degree distribution.}
% The degree of node \( i \) is \( d_i = \sum_j A_{ij} \). The empirical degree distribution is
% \begin{equation}
% P(k) = \frac{|\{i \in V : d_i = k\}|}{n}.
% \end{equation}
% Practically, $P(k)$ is one of the fastest sanity-checks for whether a synthetic graph resembles the intended regime
% (e.g., near-Poisson for Erd\H{o}s--R\'enyi versus heavy-tailed/hub-dominated graphs) \citep{drobyshevskiy_random_2020}.

\textbf{Degree inequality (Gini coefficient).}
Given node degrees sorted as $d_{1}\le \cdots \le d_{n}$ for $n$ nodes in $V_t$ with mean $\bar d$,
the definition of the gini coefficient is
\begin{equation}
c_\mathrm{gini}
= \frac{1}{2 n^2 \bar{d}}
\sum_{i=1}^{n} \sum_{j=1}^{n} |d_i - d_j|.
\end{equation}
In applications, a $c_{\mathrm{gini}}$ helps quantify hub dominance, load balancing concerns,
and whether a model produces unrealistically concentrated connectivity.

\textbf{Triangle count.}
The number of triangles is computed as
\begin{equation}
T = \frac{\mathrm{tr}(A^3)}{6}.
\end{equation}
The triangle count is a key indicator of local closure and community structure. Generators often match triangle statistics less accurately than degree distributions, exposing mode collapse.

\textbf{Global clustering coefficient.}
The total number of connected triplets is
\begin{equation}
\mathrm{Triplets}
= \sum_{i=1}^{n} \binom{d_i}{2}
= \sum_{i=1}^{n} \frac{d_i (d_i - 1)}{2}.
\end{equation}
The clustering coefficient is then
\begin{equation}
c_\mathrm{cluster} = \frac{3T}{\mathrm{Triplets}}.
\end{equation}
This captures how frequently a node’s neighbors are themselves connected. It differentiates real-world networks from random baselines and is critical for evaluating realism in graph generation. 

\textbf{Spectral gap (or algebraic connectivity).}
Let \( L = I - D^{-1/2} A D^{-1/2} \) be the normalized Laplacian. Then, the spectral gap is given by the second eigenvalue of $L$
\begin{equation}
\lambda_2(L) \in [0,2].
\end{equation}
Higher values indicate strong connectivity, robustness, and fast mixing. This metric is often maximized in resilient network design.

% \textbf{Assortativity.}
% Let
% \begin{equation}
% \mu = \frac{1}{2m} \sum_{(u,v) \in E} (d_u + d_v).
% \end{equation}
% The Pearson degree assortativity coefficient is
% \begin{equation}
% r
% = \frac{\sum_{(u,v) \in E} (d_u - \mu)(d_v - \mu)}
% {\sum_{(u,v) \in E} (d_u - \mu)^2}.
% \end{equation}
% Assortativity distinguishes between social (typically assortative) and technological/biological (often disassortative) networks, affecting diffusion and resilience \citep{newman_assortative_2002}

\begin{table}[h]
    \caption{Graph- and drawing-level metrics used for evaluation.}
    \label{tab:metrics_summary}
    \vspace{0.35em}
    \centering
    \begin{tabular}{@{}lll@{}}
        \toprule
            \textbf{Metric} & \textbf{Input} & \textbf{Typical Use} \\ \midrule
            Gini coefficient \( c_\mathrm{gini} \) & adjacency & hub dominance / inequality \\
            Triangles \( T = \mathrm{tr}(A^3)/6 \) & adjacency & closure / community signal \\
            Clustering \( c_\mathrm{cluster} = 3T/\mathrm{Triplets} \) & adjacency & local redundancy \\
            Algebraic connectivity \( \lambda_2(L) \) & adjacency & connectivity strength \\
            Angular resolution \( \mathrm{AR} \) & drawing angles & readability at vertices \\
            Edge uniformity score \( S_\ell \) & drawing lengths & regularity / aesthetics \\
            Circularity \( r_\mathrm{iso} \) & boundary polygon & roundness of layout \\
            Squareness \( r_\mathrm{square} \) & boundary polygon & alignment with square shape \\
        \bottomrule
    \end{tabular}
\end{table}

\subsection{Geometric Metrics}

The geometric metrics assume a planar drawing with node positions
\( y_i \in \mathbb{R}^2 \).
The described planar graph metrics are based on \citep{mooney_multi-dimensional_2024, mooney_universal_2025, purchase_metrics_2002}

\textbf{Angular resolution.} This metric is central in graph drawing aesthetics, where evenly spaced edge orientations improve visual clarity. Angular resolution $\mathrm{AR}$ also correlates with mesh quality in finite-element analysis.
For any node \( i \) in $V_t$ with degree \( d_i \ge 2 \), let \( \alpha_{\min}(i) \) be the minimum angle between incident edges. First, the ideal angle is defined as
\begin{equation}
\alpha_{\mathrm{ideal}}(i) = \frac{2\pi}{d_i}.
\end{equation}
Then, the per-node angular resolution score is
\begin{equation}
\mathrm{AR}(i)
= 1 - \frac{|\alpha_{\mathrm{ideal}}(i) - \alpha_{\min}(i)|}
{\alpha_{\mathrm{ideal}}(i)}.
\end{equation}
The global angular resolution is
\begin{equation}
\mathrm{AR}
= \frac{1}{|\{i : d_i \ge 2\}|}
\sum_{i : d_i \ge 2} \mathrm{AR}(i).
\end{equation}

\textbf{Edge-length uniformity.}
Let \( \ell_e \) be the Euclidean length of edge \( e \) and \( \bar{\ell} \) the mean edge length. We define
\begin{equation}
\Delta_\ell
= \frac{1}{m} \sum_{e \in E}
\frac{|\ell_e - \bar{\ell}|}{\bar{\ell}},
\qquad
S_\ell = \frac{1}{1 + \Delta_\ell},
\end{equation}
with $S_\ell$ representing the edge-length score metric, where a higher score indicates better edge-length uniformity.
Uniform edge lengths enhance readability and fairness in graph layouts. This metric supports assessments of isotropic meshing or layout aesthetics, and is often jointly optimized with angular resolution.

% \textbf{Node uniformity.}
% Let \( d_{ij} = \|p_i - p_j\|_2 \) for \( i \ne j \). The node uniformity score is
% \begin{equation}
% U_V = \frac{\min_{i \ne j} d_{ij}}
% {\max_{i \ne j} d_{ij}}.
% \end{equation}
% Node uniformity reflects spacing consistency. Low values suggest crowding or excessive spread, affecting graph readability and layout fairness. This metric helps identify overly dense or sparse regions in a drawing

\textbf{Circularity.}
Recalling that \( V_t^b \subseteq V_t \) denote boundary nodes, let \( A_b \) be the area enclosed by the polygon formed by the boundary nodes, and \( P_b \) its perimeter. The circularity score is defined as the isoperimetric ratio
\begin{equation}
r_\mathrm{iso} = \frac{4\pi A_b}{P_b^2}.
\end{equation}
This value is maximized (equal to 1) for a perfect circle of the graph boundary hull and decreases with elongation or irregularity of the boundary shape. For all baseline methods, the notion of boundary nodes does not exist, hence we use alpha shapes \cite{edelsbrunner_shape_1983} to estimate the concave hull of baseline graphs. The value of $\alpha$ is estimated using bisection, with a maximum value $\alpha=5$ and a maximum of 10 bisection steps. The bisection selects the largest $\alpha$ for which all nodes are still included in the concave hull and the hull forms a single connected polygonal domain.

\textbf{Squareness.}
To compute squareness of a graphs' hull, we consider all possible planar rotations of the boundary region and identify the orientation that minimizes the area of the axis-aligned bounding square. Let \( A_\mathrm{box} \) be the area of this minimum-area enclosing square. The squareness coefficient is
\begin{equation}
r_\mathrm{square} = \frac{A_b}{A_\mathrm{box}}.
\end{equation}
This value lies in the interval \( [0,1] \), with higher values indicating that the boundary closely conforms to a square, up to rotation. This metric is rotationally invariant and captures alignment with square layouts.

\section{Graph Generators}
\label{app:graph_gen_benchmark}

We evaluate graph algorithms using synthetic datasets generated from well-established random graph models, focusing on classical, planar, and diffusion-based generators.
Additionally, we include the House of Graphs (HOG) \cite{brinkmann_house_2013} dataset in the benchmarks, which includes, to the best of our knowledge, the most comprehensive connected planar graph dataset, albeit drawn without an embedding.
% Graphs are converted to \textsc{NetworkX} format for downstream metric computation and layout rendering.
Across all generators, benchmark-favorable settings and method-specific adaptations described throughout this chapter are applied to align each approach with the benchmark objectives. While computational budgets are kept as comparable as possible, diffusion-based models still require substantially higher training time. For all methods that do not provide an embedding by default, we use the force-directed spring layout to pre-process the graph to be embedded \cite{spring_layout} using the \textsc{NetworkX} library \cite{hagberg_exploring_2008}. This library is also used in the following described graph generators.

\subsection{Classical Random Graph Models}

\textbf{Erdős–Rényi (ER) Graphs.}
We use the $G(n,p)$ model, where each edge between the $n$ nodes is independently included with probability \( p = \frac{2m}{n(n-1)} \). This model serves as a uniform random baseline and produces graphs with a binomial degree distribution, implemented using \texttt{networkx.erdos\_renyi\_graph} \citep{Erdos_1959_pmd}. 

For benchmarking, we have generated 10{,}000 graphs per configuration, with edge probabilities ranging from 0.04 to 0.10. Connectivity is enforced, and disconnected graphs are filtered out. Higher edge densities substantially reduce the proportion of planar graphs, reflecting the well-known trade-off between connectivity and planarity in sparse regimes. Specifically, this method does not produce any planar graph for the second and third benchmark, see Appendix \ref{app:extended_results}. Despite extensive exploration of this parameter space, Erdős--Rényi graphs never achieve the best performance in our evaluations. This behavior is expected because the model is constraint-free and not goal-directed.

%\textbf{Barabási–Albert (BA) Graphs.}
%BA graphs are generated by preferential attachment, starting from a small seed graph and incrementally adding nodes with $m$ edges each. Each new edge connects to existing nodes with probability proportional to node degree, implemented using \texttt{networkx.barabasi\_albert\_graph} \citep{barbasi_albert}. We vary the number of attachments from 1 to 4, generating 40,000 instances per setting. Lower attachment values lead to sparse and often disconnected graphs, which are rejected to ensure connectivity; higher values produced dense, hub-dominated structures, all of which were non-planar.

\textbf{Exponential Random Graph Models (ERGM).}
Graphs are sampled from an exponential distribution over structural features using a Metropolis–Hastings MCMC sampler:
\begin{equation}
    P(G) \propto \exp\left(\beta_{edge} \cdot \#\text{edges} + \beta_{triangle} \cdot \#\text{triangles} + \beta_{star} \cdot \#\text{2-stars}\right).
\end{equation}
We use the implementation from \citep{snijders_markov_2002} performing 1000 MCMC steps after burn-in. Parameters are selected from $\beta_{\text{edge}} \in \{-2.5, -2.0, -1.5\}$ and $\beta_{\text{triangle}} \in \{0.0, 0.1\}$ with fixed $\beta_{\text{star}} = 0.0$. Compared to other models, ERGMs involve a large hyperparameter space and are more difficult to tune, as certain parameter settings lead to disconnected graphs, while others produce dense, non-planar structures. As ERGMs do not enforce planarity, many sampled graphs violate planarity.

\subsection{Planar Graph Generators}

\textbf{Plantri.}
We generate triangulated planar graphs using the \textsc{Plantri} software \citep{brinkmann_fast_2007}, which guarantees planarity and non-redundancy via combinatorial generation and isomorphism rejection. Due to expensive I/O and limited scaling beyond 15–20 nodes, we evaluate Plantri only on the first 2{,}000 sampled graphs.

\textbf{Boltzmann Sampler.}
Uniform planar graphs are sampled using a Boltzmann sampler based on analytic combinatorics \citep{fusy_uniform_2009}, implemented via the \texttt{planar\_graph\_sampler} library \citep{towink_boltzmann_2020}. This approach generates unbiased planar graphs with expected quadratic time complexity. Due to the relatively high computational cost of this approach, we evaluate 50 graphs per benchmark, as larger sample sizes are a lot more expensive compared to our method.

\textbf{Delaunay Triangulations.}
A set of \(|G|-b>0\) points are sampled uniformly from circular and square domains, while an additional \(b\) boundary points placed along the domain boundary \textit{equidistantly}. The resulting point set is triangulated according to the Delaunay criterion using \texttt{scipy.spatial.Delaunay}. For circular target metrics, boundary points are sampled along a circle to enforce a larger isoperimetric ratio, while an analogous construction is used for the squareness metric.

\textbf{Lattice Graphs.}
Planar grid graphs are generated as regular 2D lattices using \texttt{networkx.grid\_2d\_graph}. For a target graph size, we select an approximately square grid with 4-neighbor connectivity; in the circular setting, rows are wrapped into rings. These graphs provide deterministic spatial baselines with highly regular structure and no triangles, although the final node count may differ slightly from the target size.

\subsection{Deep Generative Graph Models}
We further include representative modern Deep Generative Models (DGM) for graph generation as baselines, including DiGress, GDSS and MuDiff \citep{vignac_digress_2023, jo2022scorebasedgenerativemodelinggraphs, hua2024mudiffunifieddiffusioncomplete}. These models are distribution-learning approaches that generate graphs by sampling from learned probabilistic processes. While DiGress and MuDiff provide a conditional sampling mechanisms towards target properties, they do not enforce hard constraints during generation. The same applies to GDSS, which is a purely unconditional model, but can be enhanced with PRODIGY \citep{pmlr-v235-sharma24b} guidance towards single-objective target properties. Due to this, all models are evaluated by selecting candidates according to the procedure described in Section~\ref{app:graph-selection-proc} to adhere to as many constraints as possible for a fair comparison.

\textbf{DiGress.} DiGress is a discrete diffusion model for categorical graph generation with optional conditional guidance \cite{vignac_digress_2023}. The model iteratively denoises graph structures in discrete space and enables conditioning through learned property predictors. In our evaluation, DiGress is first trained once in an unconditional manner on the full HOG dataset to learn a general distribution over connected graphs. Subsequently, for each benchmark, we train a dedicated property regressor that estimates the respective benchmark-specific objective (e.g., number of triangles and isoperimetric ratio in Benchmark \ref{app:benchmark_4}) which is trained using a per-metric standardized mean-squared error loss, so that metrics on different scales contribute comparably to the objective. This regressor is then used within the conditional generation mechanism of DiGress to guide sampling toward graphs that optimize the target objective. 

All architectural settings follow the original DiGress implementation. Minor adjustments to the number of diffusion steps and training epochs were applied to ensure stable convergence on the HOG dataset. More detailed hyperparameters are shown in Table~\ref{tab:digress_hparams}.

We also note that the original DiGress planar benchmark was evaluated on the SPECTRE planar dataset, consisting of 200 graphs of fixed size (64 nodes) generated exclusively as Delaunay triangulations. This dataset exhibits limited structural diversity compared to HOG, which includes 4000 graphs of variable size (up to 250 nodes) and unrestricted topology. Under this more challenging regime, only a subset of unconditionally generated samples satisfies the planarity constraint (33\%), while the original work of \citet{vignac_digress_2023} report 75\% validity of planar graphs in the simpler dataset regime used.

\begin{table}[H]
\caption{Hyperparameters for DiGress with property-regressor guidance.}
\vspace{0.35em}
\label{tab:digress_hparams}

    \centering
    \begin{tabular}{ll}
        \toprule
        \textbf{Hyperparameter} & \textbf{Value} \\
        \midrule    
        Unconditional epochs & 1000 \\
        Unconditional batch size & 8 \\
        Unconditional learning rate & $2 \cdot 10^{-4}$ \\
        Diffusion steps & 500 \\
        Regressor hidden dimension & 128 \\
        Regressor dropout & 0.1 \\
        Regressor epochs & 200 \\
        Regressor batch size & 32 \\
        Regressor learning rate & $10^{-3}$ \\
        Regressor weight decay & $10^{-5}$ \\
        Regressor scheduler & CosineAnnealingLR \\
        Inference Guidance scale & $2.0$ \\
        \bottomrule
    \end{tabular}
\end{table}

\textbf{GDSS with PRODIGY Guidance.} GDSS is a score-based generative model that learns continuous diffusion dynamics over adjacency matrices and node features \cite{jo2022scorebasedgenerativemodelinggraphs}. In our evaluation, an unconditional GDSS model is first trained on the HOG dataset to learn a general distribution over graph structures. Because GDSS does not natively support conditional generation, we apply guidance using the PRODIGY framework \cite{pmlr-v235-sharma24b}, which steers sampling toward specified single-objective target properties. We select the target node number as the conditioning signal during graph generation with Prodigy. More detailed hyperparameters are shown in Table~\ref{tab:gdss_prodigy_hparams} to enhance reproducibility.

\begin{table}[t]
\caption{Hyperparameters for GDSS with PRODIGY guidance.}
\vspace{0.35em}
\label{tab:gdss_prodigy_hparams}
    \centering
    \begin{tabular}{ll}
        \toprule
        \textbf{Hyperparameter} & \textbf{Value} \\
        \midrule
            Training epochs & 2000 \\
            Batch size & 64 \\
            Learning rate & $10^{-2}$ \\
            EMA decay & 0.999 \\
            Diffusion steps & 1000 \\
            Predictor / Corrector & Euler / Langevin \\
        \bottomrule
    \end{tabular}
\end{table}

\textbf{MuDiff.} MuDiff is a diffusion-based model originally developed for joint structure–geometry generation in molecular graphs \cite{hua2024mudiffunifieddiffusioncomplete}. Although designed for three-dimensional molecular systems, the architecture can be adapted to general graph generation tasks. For compatibility with our planar graph setting, we preprocess HOG graphs using force-directed spring layout and embed them in three-dimensional space by assigning zero values to the third coordinate. Benchmark-specific conditional models are trained following the original architectural specifications. Accordingly, four conditional models are trained independently, one for each benchmark scenario. Hyperparameters are tuned separately for each benchmark, in Table~\ref{tab:mudiff_hparams} we report the ranges. We further note that, unlike the other deep generative methods considered in this study, MuDiff directly generates graphs together with their geometric embeddings, rather than requiring a separate layout or embedding procedure, making it a relevant comparison to state of the art conditional graph generation methods in unconstrained settings.

\begin{table}[h]
\caption{Hyperparameters for MuDiff.}
\vspace{0.35em}
    \centering
    \begin{tabular}{ll}
        \toprule
        \textbf{Hyperparameter} & \textbf{Value} \\
        \midrule
        Diffusion steps & 500 \\
        Noise schedule & \texttt{polynomial\_2} \\
        Diffusion loss & \texttt{l2} \\
        Learning rate & [$5 \cdot 10^{-4}, 5 \cdot 10^{-5}$] \\
        Batch size & [2,4] \\
        Training epochs & [400,600] \\
        EMA decay & 0.999 \\
        Encoder layers & 6 \\
        Embedding dimension & 128 \\
        Edge embedding dimension & 128 \\
        Graph embedding dimension & 32 \\
        Attention heads & 8 \\
        \bottomrule
    \end{tabular}
    \label{tab:mudiff_hparams}
\end{table}

\section{Ranking-Based Graph Selection for Baseline Evaluations}
\label{app:graph-selection-proc}
We consider a set of candidate graphs ${\tilde{G}_i}$ produced by a given method. Candidates are first filtered for planarity (this can still contain graphs with edge intersections). Only planar graphs are retained for further processing through the benchmark pipeline, resulting in the set of graphs ${G_i}$.

For each retained graph, we compute benchmark metrics $\mathcal{M}(G_i)$ and the final reward $r_i$(see Section~\ref{sec:reward}). The additional quantities used for ranking are split according to the generator class and are described separately for classical and diffusion-based methods below.

\paragraph{Classical Generators}
For classical generators, we randomly sample candidates according to the generator specifications and report metrics, and constraint satisfaction. For each generator, 
we retain only the graphs that satisfy most constraints. If multiple graphs satisfy equally many constraints, we report the graph with highest reward among them. As visible in the benchmarks (cf. Appendix~\ref{app:extended_results}), classical generators rarely satisfy all constraints.

\paragraph{Deep Generative Models}

For deep generative models, we sample approximately \(M \approx 1000\) candidate graphs per generator. Since generated samples may contain small disconnected components, we retain only the largest connected component \(G_i\) with size \(|G_i|\) for subsequent evaluation, ensuring that otherwise valid planar structures are not discarded. Graph-size proximity is prioritized to ensure comparability with the target benchmark configuration: since DGMs do not explicitly control the number of nodes during generation, samples can vary substantially in size, and differences in graph size can strongly affect both evaluation metrics and reward. We therefore first rank candidates by their distance to the target size. Let $|G^\star|$ denote the target size and define $d_i = \big|\,|G_i| - |G^\star|\,\big|$ and

\begin{equation}
    b_i =
    \begin{cases}
    0, & d_i \leq 10, \\
    1 + \left\lfloor \frac{d_i - 10}{10} \right\rfloor, & \text{otherwise}.
    \end{cases}
\end{equation}

The smaller $b_i$ is for a graph, the higher the graph is ranked. We further compute the force-directed spring layout embedding (omitted for MuDiff) and evaluate the constraints satisfaction. Constraint statistics are recorded but not used for ranking, as prioritizing them was empirically found to strongly disadvantage these models compared to other generators.

The final selected graph is the highest-scoring candidate among those with the closest graph-size match to the target benchmark configuration, and is reported together with its constraint statistics.

\section{Extended Results on Benchmarks}
\label{app:extended_results}
\label{app:extended_tables}
\begin{figure}[h!]
  \begin{center}
    \centerline{\includegraphics[width=1\columnwidth]{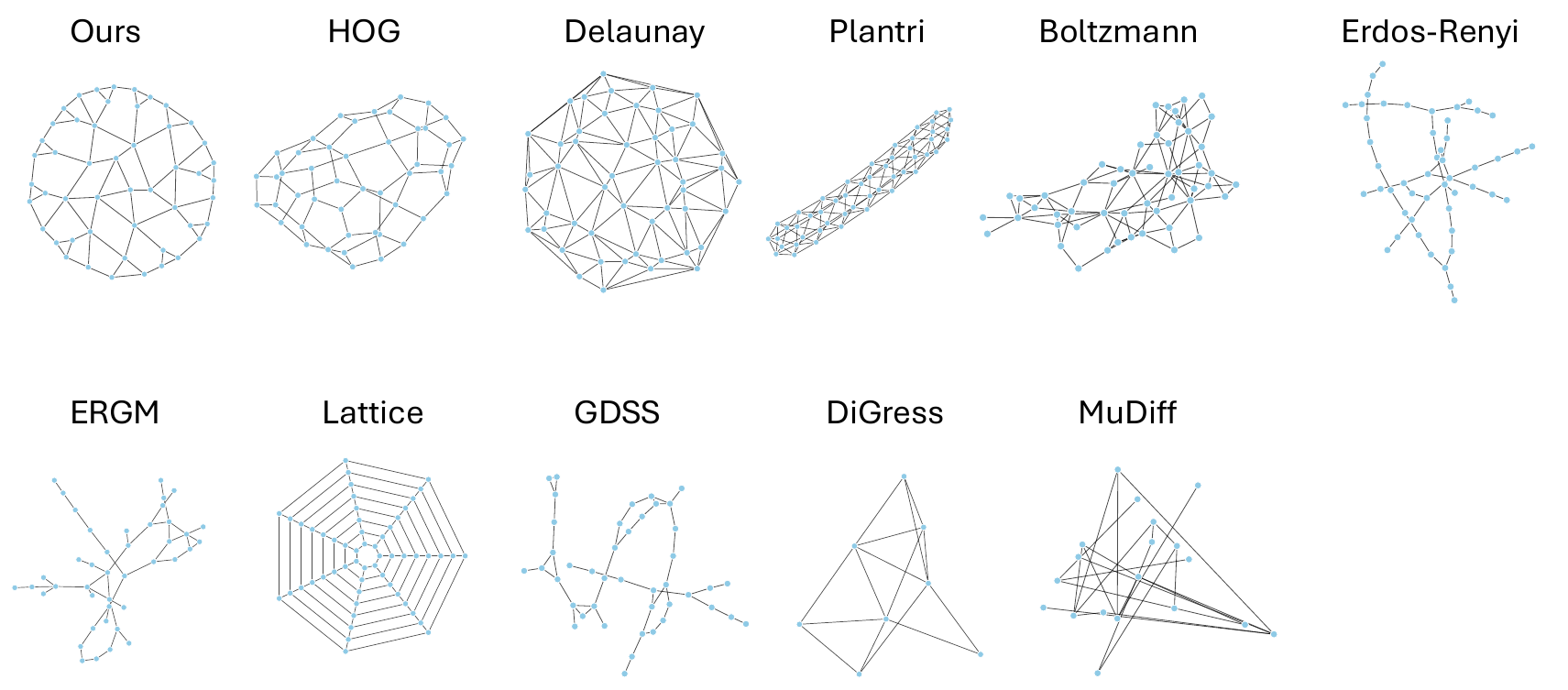}}
    \caption{Results for the power grid case for our method and all baselines.}
    \label{fig:comparison_case1}
  \end{center}
\end{figure}

\begin{figure}[h!]
  \begin{center}
    \centerline{\includegraphics[width=1\columnwidth]{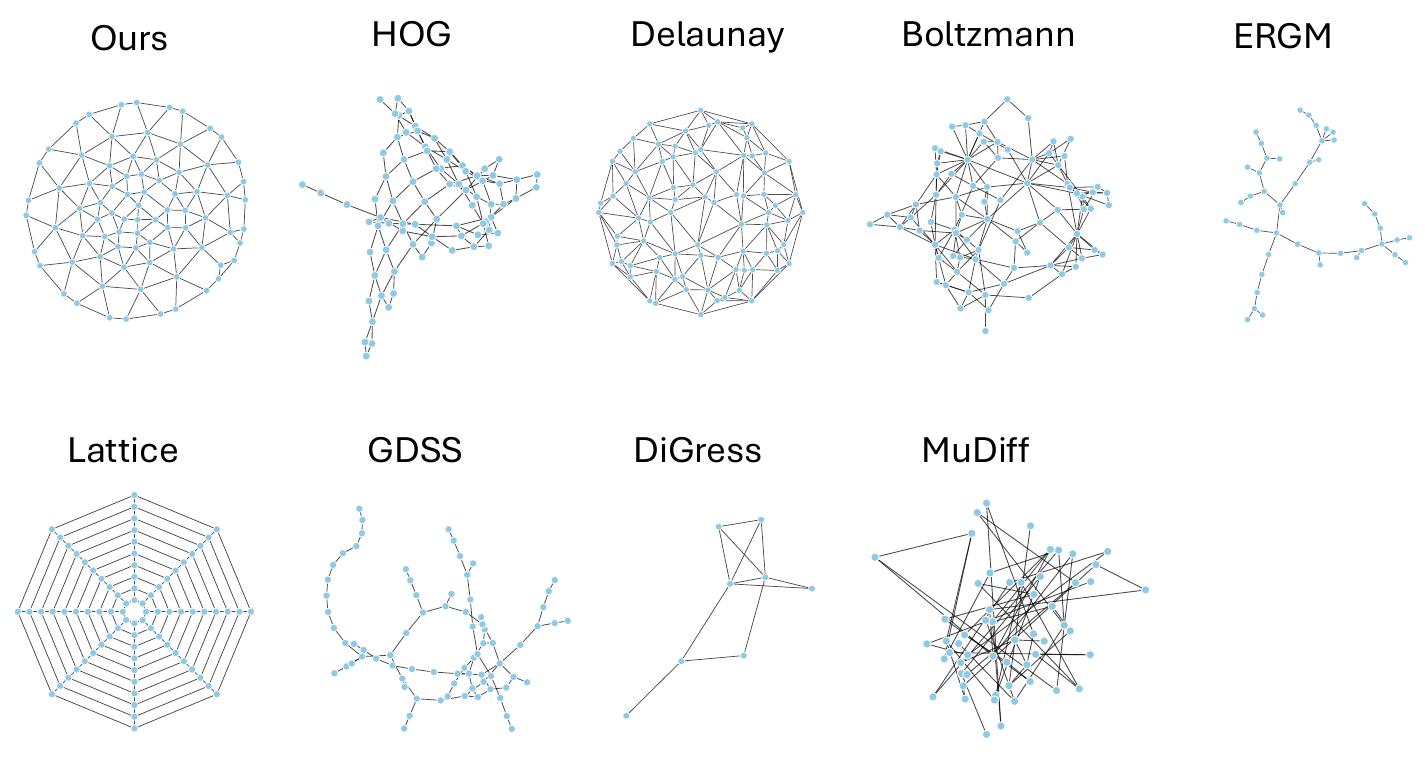}}
    \caption{Results for the urban road grid case for our method and all baselines.}
    \label{fig:comparison_case2}
  \end{center}
\end{figure}

\begin{figure}[h!]
  \begin{center}
    \centerline{\includegraphics[width=1\columnwidth]{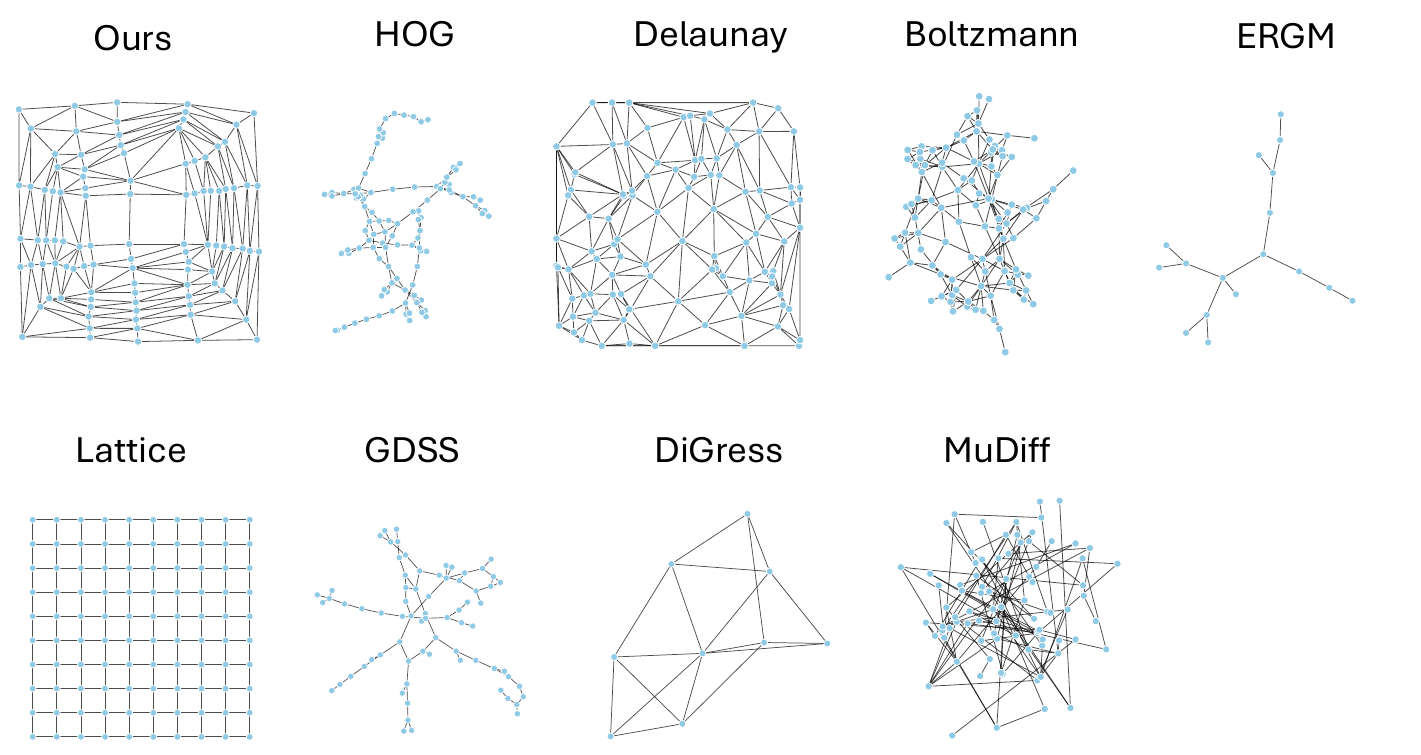}}
    \caption{Results for the mesh in square hull high spectral gap for our method and all baselines.}
    \label{fig:comparison_case3}
  \end{center}
\end{figure}

\begin{figure}[h]
  \begin{center}
    \centerline{\includegraphics[width=1\columnwidth]{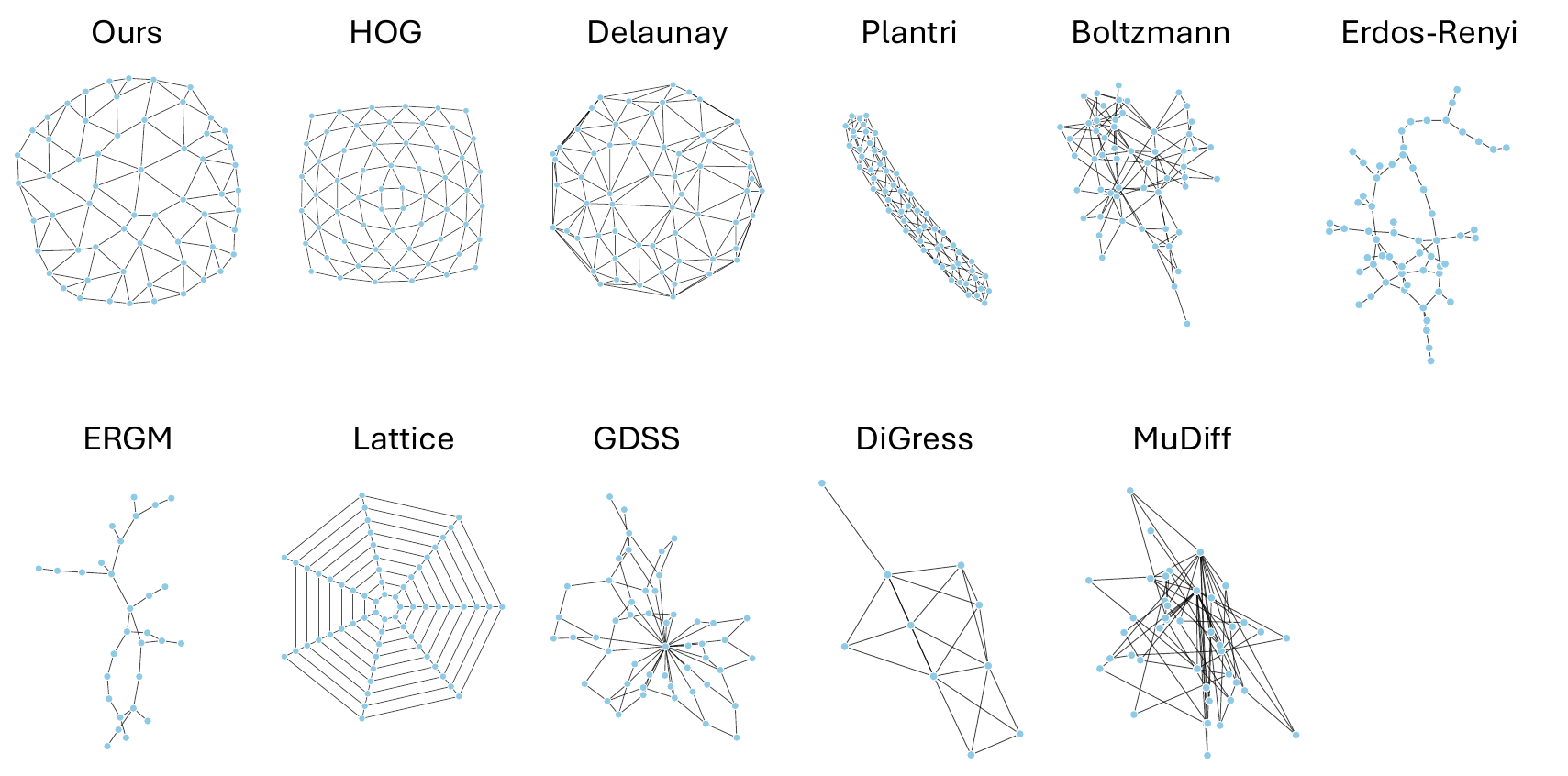}}
    \caption{Results for the graph in a circular hull with 40 triangles case for our method and all baselines.}
    \label{fig:comparison_case4}
  \end{center}
\end{figure}

This section reports the full benchmark results (Tables~\ref{tab:extended-power-grid-circular}–\ref{tab:extended-mesh-with-triangles}) and complements them with additional qualitative comparisons (Figures~\ref{fig:comparison_case1}–\ref{fig:comparison_case4}).

Some benchmark families, such as Delaunay triangulations and lattice grids, provide explicit planar embeddings by construction. For generators without intrinsic geometric embeddings, including the HOG dataset, node coordinates for visualization and geometric evaluation are computed using a force-directed spring layout \citep{spring_layout}.
Each table presents the overall reward (cf. Equation~\ref{eq:reward}), which quantifies the multi-objective optimization score, together with the individual  metrics and constraint satisfaction for all baselines, including both conventional algorithms and deep generative models. The constraint $\ell_1$ denotes whether all edge lengths in a graph satisfy $\ell\in[0.75,2.5]$, where for $\ell_2$ the interval is $[0.1,3.5]$. We additionally include 1-sigma standard deviations for our method in all tables, as well as constraint satisfaction for each baseline. We further denote with \textit{Emb}., whether a baseline produces an embedding by default, or if the graph is post-processed using the force-directed spring layout.
Table~\ref{tab:all-benchmark-deviations} additionally reports the per-metric relative deviations across all four benchmarks, marking each metrics' contribution to the aggregate reward explicit (Equation~\ref{eq:reward}).

\clearpage
\subsection{Benchmark 1: Power Grid in Circular Hull}
\begin{table}[h!]
  \caption{Extended results for power grid in circular hull with 50 nodes. Note that although MuDiff achieves highest spectral gap, this is achieved by not respecting planarity constraints and target graph size. Generally, for smaller graph the spectral gap tends to be higher, as well as for non-planar graphs.}
  \vspace{0.35em}
  \label{tab:extended-power-grid-circular}
  \centering
  \setlength{\tabcolsep}{3.0pt}
  \renewcommand{\arraystretch}{1.0}

  \begin{tabular}{lccccc|cccccc}
    \toprule
    & \multicolumn{1}{c}{Reward}
    & \multicolumn{1}{c}{Geom.}
    & \multicolumn{2}{c}{Topological}
    & \multicolumn{1}{c|}{Spec.}
    & \multicolumn{6}{c}{Constraints} \\
    \cmidrule(lr){2-2}
    \cmidrule(lr){3-3}
    \cmidrule(lr){4-5}
    \cmidrule(lr){6-6}
    \cmidrule(l){7-12}
    Method
    & $r$
    & $r_{\mathrm{iso}}$
    & $c_{\mathrm{gini}}$
    & $c_{\mathrm{clust}}$
    & $\lambda_2(L)$
    & $|G|$
    & $\alpha_{\min}$
    & $c_{\ell}$
    & $\deg_{\max}$
    & $c_{\mathrm{pl}}$
    & Emb. \\
    \midrule

    Target
      & 1.0
      & 1.0 & 0.1 & 0.2 & 1.0
      & 50 & $20^{\circ}$ & $\ell_1$ & 8 & True & True \\

    \midrule

    \multirow{2}{*}{\textbf{Ours}}
      & \textbf{0.756}
      & \textbf{0.960} & \textbf{0.090} & \textbf{0.200} & 0.050
      & \multirow{2}{*}{\textbf{50}}
      & \multirow{2}{*}{$\checkmark$}
      & \multirow{2}{*}{$\checkmark$}
      & \multirow{2}{*}{$\checkmark$}
      & \multirow{2}{*}{$\checkmark$}
      & \multirow{2}{*}{$\checkmark$} \\
      & $\pm 0.004$
      & $\pm 0.020$ & $\pm 0.002$ & $\pm 0.004$ & $\pm 0.003$
      & & & & & & \\

    \midrule

    HOG
      & 0.382
      & 0.841 & 0.031 & 0.000 & 0.075
      & 47 & $\checkmark$ & $\checkmark$ & $\checkmark$ & \ding{55} & \ding{55}  \\

    Delaunay
  & 0.698

  & 0.932 & 0.182 & 0.217 & \textbf{0.087}

  & \textbf{50} & \ding{55} & \ding{55} & $\checkmark$ & $\checkmark$ & $\checkmark$ \\

    Plantri
      & 0.445
      & 0.377 & 0.034 & 0.417 & 0.039
      & \textbf{50} & \ding{55} & $\checkmark$ & $\checkmark$ & \ding{55} & \ding{55} \\

    Boltzmann
      & 0.545
      & 0.468 & 0.279 & 0.248 & 0.064
      & 52 & \ding{55} & \ding{55} & \ding{55} & \ding{55} & \ding{55} \\

    Erd\H{o}s--R\'enyi
      & 0.467
      & 0.697 & 0.195 & 0.043 & 0.006
      & \textbf{50} & $\checkmark$ & $\checkmark$ & $\checkmark$ & \ding{55} & \ding{55} \\

    ERGM
      & 0.526
      & 0.669 & 0.272 & 0.114 & 0.020
      & 42 & $\checkmark$ & $\checkmark$ & $\checkmark$ & \ding{55} & \ding{55} \\

    Lattice
      & 0.431
      & 0.932 & 0.051 & 0.000 & 0.043
      & 56 & $\checkmark$ & \ding{55} & $\checkmark$ & $\checkmark$ & $\checkmark$ \\

    \midrule

    DiGress
      & 0.671
      & 0.458 & 0.181 & 0.456 & 0.579
      & 8 & \ding{55} & $\checkmark$ & $\checkmark$ &  \ding{55}  & \ding{55} \\

    GDSS-Prodigy
      & 0.557
      & 0.495 & 0.2 & 0.16 & 0.006
      & 43 & $\checkmark$ & \ding{55} & $\checkmark$ & \ding{55} & \ding{55} \\

    MuDiff
      &  0.520
      & 0.140 & 0.496 & 0.516 & 0.889
      & 19 & \ding{55} & \ding{55} &  $\checkmark$ & \ding{55}& $\checkmark$ \\

    \bottomrule
  \end{tabular}
  \vspace{-0.1in}
\end{table}

\clearpage

\subsection{Benchmark 2: Urban Road Grid in Circular Hull}

\begin{table}[h!]
  \caption{Extended result for urban road grid in circular hull with 80 nodes. Note that although some baselines marginally improve individual metrics, the combined multi-objective reward is highest for our method that balances the individual metrics closest to the target on average. Additionally, the full constraint set is not respected by most baselines, e.g. ERGM beats the angular resolution metric, but does not satisfy target graph size.}
  \vspace{0.35em}
  \label{tab:urban-road-circular}
  \centering
  \setlength{\tabcolsep}{3.0pt}
  \renewcommand{\arraystretch}{1.0}

  \begin{tabular}{lccccc|cccccc}
    \toprule
    & \multicolumn{1}{c}{Reward}
    & \multicolumn{3}{c}{Geometric}
    & \multicolumn{1}{c|}{Topo.}
    & \multicolumn{6}{c}{Constraints} \\
    \cmidrule(lr){2-2}
    \cmidrule(lr){3-5}
    \cmidrule(lr){6-6}
    \cmidrule(l){7-12}
    Method
    & $r$
    & $r_{\mathrm{iso}}$
    & $S_l$
    & $\mathrm{AR}$
    & $c_{\mathrm{gini}}$
    & $|G|$
    & $\alpha_{\min}$
    & $c_{\ell}$
    & $\deg_{\max}$
    & $c_{\mathrm{pl}}$
    & Emb. \\
    \midrule

    Target
      & 1.0
      & 1.0 & 0.7 & 0.8 & 0.2
      & 80 & $20^{\circ}$ & $\ell_1$ & 8 & True & True \\

    \midrule

    \multirow{2}{*}{\textbf{Ours}}
      & \textbf{0.942}
      & \textbf{0.952} & 0.82 & \textbf{0.755} & 0.163
      & \multirow{2}{*}{\textbf{80}}
      & \multirow{2}{*}{$\checkmark$}
      & \multirow{2}{*}{$\checkmark$}
      & \multirow{2}{*}{$\checkmark$}
      & \multirow{2}{*}{$\checkmark$}
      & \multirow{2}{*}{$\checkmark$} \\
      & $\pm 0.011$
      & $\pm 0.027$ & $\pm 0.02$ & $\pm 0.019$ & $\pm 0.01$
      & & & & & & \\

    \midrule

    HOG
      & 0.782
      & 0.403 & 0.832 & 0.486 & 0.158
      & \textbf{80} & \ding{55} & \ding{55} & $\checkmark$ & \ding{55} & \ding{55} \\

Delaunay

  & 0.914

  & \textbf{0.952} & \textbf{0.758} & 0.445 & \textbf{0.200}

  & \textbf{80} & \ding{55} & \ding{55} & $\checkmark$ & $\checkmark$ & $\checkmark$ \\

    Plantri
      & --
      & -- & -- & -- & --
      & -- & -- & -- & -- & -- & \ding{55} \\

    Boltzmann
      & 0.798
      & 0.649 & 0.772 & 0.364 & 0.282
      & 81 & \ding{55} & \ding{55} & \ding{55} & \ding{55} & \ding{55} \\

    Erd\H{o}s--R\'enyi
      & --
      & -- & -- & -- & --
      & -- & -- & -- & -- & -- & \ding{55} \\

    ERGM
      & 0.874
      & 0.574 & 0.797 & \textbf{0.816} & 0.276
      & 45 & $\checkmark$ & $\checkmark$ & $\checkmark$ & $\checkmark$ & \ding{55} \\

    Lattice
      & 0.799
      & 0.948 & 0.578 & 0.750 & 0.043
      & \textbf{80} & $\checkmark$ & \ding{55} & $\checkmark$ & $\checkmark$ & $\checkmark$ \\

    \midrule

    DiGress
      & 0.751
      & 0.479 & 0.823 & 0.336 & 0.274
      & 8 & \ding{55} & $\checkmark$ & $\checkmark$ & \ding{55} & \ding{55} \\

    GDSS-Prodigy
      & 0.834
      & 0.291 & 0.802 & \textbf{0.790} & 0.185
      & 76 & \ding{55}  & \ding{55} & $\checkmark$ & \ding{55} & \ding{55} \\
      
    MuDiff
      & 0.534
      & 0.338 & 0.006 & 0.462 & 0.254
      & 56 & \ding{55} & \ding{55} &  $\checkmark$ & \ding{55}& $\checkmark$  \\

    \bottomrule
  \end{tabular}
\end{table}

\clearpage

\subsection{Benchmark 3: Strongly Connected Mesh in Square Hull}
\label{app:benchmark_3}
%% TABLE 3
\begin{table}[h!]
  \caption{Extended results for highly connected mesh in square hull with 100 nodes. This benchmark relaxes the constraint set compared to the other benchmarks, providing insights to graph generation closer to an unconstrained setting. Note that in this benchmark, MuDiff achieves highest spectral gap, but does so by strongly dissatisfying planarity constraints.}
  \label{tab:extended-electrical-mesh-rect}
  \vspace{0.35em}
  \centering
  \setlength{\tabcolsep}{3.0pt}
  \renewcommand{\arraystretch}{1.0}

  \begin{tabular}{lccc|cccccc}
    \toprule
    & \multicolumn{1}{c}{Reward}
    & \multicolumn{1}{c}{Geometric}
    & \multicolumn{1}{c|}{Spectral}
    & \multicolumn{6}{c}{Constraints} \\
    \cmidrule(lr){2-2}
    \cmidrule(lr){3-3}
    \cmidrule(lr){4-4}
    \cmidrule(l){5-10}
    Method
    & $r$
    & $r_{\mathrm{square}}$
    & $\lambda_2(L)$
    & $|G|$
    & $\alpha_{\min}$
    & $c_{\ell}$
    & $\deg_{\max}$
    & $c_{\mathrm{pl}}$
    & Emb. \\
    \midrule

    Target
      & 1.0
      & 1.0 & 0.300
      & 100 & $1^{\circ}$ & $\ell_2$ & 15 & True & True \\

    \midrule

    \multirow{2}{*}{\textbf{Ours}}
      & \textbf{0.647}
      & 0.887 & \textbf{0.064}
      & \multirow{2}{*}{\textbf{100}}
      & \multirow{2}{*}{$\checkmark$}
      & \multirow{2}{*}{$\checkmark$}
      & \multirow{2}{*}{$\checkmark$}
      & \multirow{2}{*}{$\checkmark$}
      & \multirow{2}{*}{$\checkmark$} \\
      & $\pm 0.01$
      & $\pm 0.06$ & $\pm 0.006$
      & & & & & & \\

    \midrule

    HOG
      & 0.360
      & 0.543 & 0.002
      & 101 & $\checkmark$ & $\checkmark$ & $\checkmark$ & \ding{55} & \ding{55} \\

    Delaunay
      & 0.578
      & 0.863 & 0.039
      & \textbf{100} & \ding{55}& $\checkmark$ & $\checkmark$ & $\checkmark$ & $\checkmark$ \\

    Plantri
      & --
      & -- & --
      & -- & -- & -- & -- & -- & \ding{55} \\

    Boltzmann
      & 0.502
      & 0.627 & 0.040
      & \textbf{100} & \ding{55} & $\checkmark$ & $\checkmark$ & \ding{55} & \ding{55} \\

    Erd\H{o}s--R\'enyi
      & --
      & -- & --
      & -- & -- & -- & -- & -- & \ding{55} \\

    ERGM
      & 0.487
      & 0.486 & 0.057
      & 17 & $\checkmark$ & $\checkmark$ & $\checkmark$ & $\checkmark$ & \ding{55} \\

    Lattice
      & 0.587
      & \textbf{1.000} & 0.029
      & \textbf{100} & $\checkmark$ & $\checkmark$ & $\checkmark$ & $\checkmark$ & $\checkmark$ \\

    \midrule

    DiGress
      & 0.699
      & 0.385 & 0.216
      & 9 & \ding{55} & $\checkmark$ & $\checkmark$ & \ding{55} & \ding{55} \\

    GDSS-Prodigy
      & 0.386
      & 0.545 & 0.01
      & 72 & $\checkmark$  &$\checkmark$ & $\checkmark$ & \ding{55} & \ding{55} \\

    MuDiff
      &  0.410
      &  0.197 & 0.918
      & 85 & \ding{55} & \ding{55} &  $\checkmark$ & \ding{55}& $\checkmark$ \\

    \bottomrule
  \end{tabular}
\end{table}

\clearpage

\subsection{Benchmark 4: Graph with 40 triangles in circular hull}
\label{app:benchmark_4}
\begin{table}[h!]
  \caption{Extended results for graph in circular hull with 60 nodes and 40 triangles.}
  \vspace{0.35em}
  \label{tab:extended-mesh-with-triangles}
  \centering
  \setlength{\tabcolsep}{3.0pt}
  \renewcommand{\arraystretch}{1.0}

  \begin{tabular}{lccc|cccccc}
    \toprule
    & \multicolumn{1}{c}{Reward}
    & \multicolumn{1}{c}{Geom.}
    & \multicolumn{1}{c|}{Topological}
    & \multicolumn{6}{c}{Constraints} \\
    \cmidrule(lr){2-2}
    \cmidrule(lr){3-3}
    \cmidrule(lr){4-4}
    \cmidrule(l){5-10}
    Method
    & $r$
    & $r_{\mathrm{iso}}$
    & $T$
    & $|G|$
    & $\alpha_{\min}$
    & $c_{\ell}$
    & $\deg_{\max}$
    & $c_{\mathrm{pl}}$
    & Emb. \\
    \midrule

    Target
      & 1.0
      & 1.0 & 40
      & 60 & $20^{\circ}$ & $\ell_1$ & 8 & True & True \\

    \midrule

    \multirow{2}{*}{\textbf{Ours}}
      & \textbf{0.979}
      & 0.920 & \textbf{39.91}
      & \multirow{2}{*}{\textbf{60}}
      & \multirow{2}{*}{$\checkmark$}
      & \multirow{2}{*}{$\checkmark$}
      & \multirow{2}{*}{$\checkmark$}
      & \multirow{2}{*}{$\checkmark$}
      & \multirow{2}{*}{$\checkmark$} \\
      & $\pm 0.022$
      & $\pm 0.080$ & $\pm 0.09$
      & & & & & & \\

    \midrule

    HOG
      & 0.856
      & 0.892 & 64
      & \textbf{60} & $\checkmark$ & $\checkmark$ & $\checkmark$ & $\checkmark$ & \ding{55} \\

Delaunay

  & 0.759

  & \textbf{0.949} & 107

  & \textbf{60} & \ding{55} & \ding{55} & $\checkmark$ & $\checkmark$ & $\checkmark$ \\

    Plantri
      & 0.494
      & 0.312 & 116
      & \textbf{60} & \ding{55} & $\checkmark$ & $\checkmark$ & \ding{55} & \ding{55} \\

    Boltzmann
      & 0.875
      & 0.633 & 38
      & 56 & \ding{55} & \ding{55} & \ding{55} & \ding{55} & \ding{55} \\

    Erd\H{o}s--R\'enyi
      & 0.415
      & 0.709 & 0
      & \textbf{60} & \ding{55} & $\checkmark$ & $\checkmark$ & \ding{55} & \ding{55} \\

    ERGM
      & 0.333
      & 0.500 & 0
      & 29 & $\checkmark$ & $\checkmark$ & $\checkmark$ & \ding{55} & \ding{55} \\

    Lattice
      & 0.482
      & 0.932 & 0
      & 63 & $\checkmark$ & \ding{55} & $\checkmark$ & \ding{55} & $\checkmark$ \\

    \midrule

    DiGress
      & 0.569
      & 0.422 & 15
      & 10 &  \ding{55} & $\checkmark$ & $\checkmark$ & \ding{55} & \ding{55} \\

    GDSS-Prodigy
      & 0.552
      & 0.473 & 12
      & 50 &  \ding{55}  &  \ding{55}  &  \ding{55}  &  \ding{55}  & \ding{55} \\

    MuDiff
      & 0.433
      & 0.173 & 16
      & 42 & \ding{55} & \ding{55} & \ding{55}  & \ding{55}  & $\checkmark$ \\

    \bottomrule
  \end{tabular}
\end{table}

\begin{table}[p]
\caption{Per-metric relative deviations $d_m$ reconstructed from the rounded metrics in Tables~\ref{tab:extended-power-grid-circular}--\ref{tab:extended-mesh-with-triangles} in addition to the absolute reward. Lower is better for the deviation metrics; each metric contributes $d_m/|\mathcal{M}|$ to the reward.}
\label{tab:all-benchmark-deviations}
\centering
\footnotesize
\setlength{\tabcolsep}{5pt}
\renewcommand{\arraystretch}{0.88}
\begin{tabular}{@{}lrrrrr@{}}
\toprule
\multicolumn{6}{l}{\textbf{Benchmark 1: Resilient Power Grid}} \\
\addlinespace[2pt]
Method &  Reward & $d_{\mathrm{iso}}$ & $d_{\mathrm{gini}}$ & $d_{\mathrm{clust}}$ & $d_{\lambda_2}$ \\
\midrule
\multirow{2}{*}{\textbf{Ours}} & \textbf{0.756} & \textbf{0.020} & \textbf{0.053} & \textbf{0.000} & 0.905 \\
 & $\pm 0.004$ & $\pm 0.010$ & $\pm 0.011$ & $\pm 0.010$ & $\pm 0.005$ \\
\midrule
HOG & $0.382$ & 0.0864 & 0.5267 & 1.0000 & 0.8605 \\
Delaunay & $0.698$ & 0.0352 & 0.2908 & 0.0408 & \textbf{0.8399} \\
Plantri & $0.445$ & 0.4524 & 0.4925 & 0.3517 & 0.9249 \\
Boltzmann & $0.545$ & 0.3624 & 0.4723 & 0.1071 & 0.8797 \\
Erd\H{o}s--R\'enyi & $0.467$ & 0.1786 & 0.3220 & 0.6461 & 0.9881 \\
ERGM & $0.526$ & 0.1983 & 0.4624 & 0.2739 & 0.9608 \\
Lattice & $0.431$ & 0.0352 & 0.3245 & 1.0000 & 0.9175 \\
\midrule
DiGress & $0.671$ & 0.3717 & 0.2883 & 0.3902 & 0.2666 \\
GDSS-Prodigy & $0.557$ & 0.3378 & 0.3333 & 0.1111 & 0.9881 \\
MuDiff & $0.520$ & 0.7544 & 0.6644 & 0.4413 & 0.0588 \\
\midrule
\multicolumn{6}{l}{\textbf{Benchmark 2: Urban Road Grid}} \\
\addlinespace[2pt]
Method &  Reward & $d_{\mathrm{iso}}$ & $d_{S_l}$ & $d_{\mathrm{AR}}$ & $d_{\mathrm{gini}}$ \\
\midrule
\multirow{2}{*}{\textbf{Ours}} & \textbf{0.942} & \textbf{0.025} & 0.079 & \textbf{0.029} & 0.102 \\
 & $\pm 0.011$ & $\pm 0.014$ & $\pm 0.012$ & $\pm 0.013$ & $\pm 0.030$ \\
\midrule
HOG & $0.782$ & 0.4255 & 0.0862 & 0.2442 & 0.1173 \\
Delaunay & $0.914$ & \textbf{0.0246} & \textbf{0.0398} & 0.2851 & \textbf{0.0000} \\
Plantri & --- & --- & --- & --- & --- \\
Boltzmann & $0.798$ & 0.2129 & 0.0489 & 0.3746 & 0.1701 \\
Erd\H{o}s--R\'enyi & --- & --- & --- & --- & --- \\
ERGM & $0.874$ & 0.2706 & 0.0648 & \textbf{0.0099} & 0.1597 \\
Lattice & $0.799$ & 0.0267 & 0.0955 & 0.0323 & 0.6461 \\
\midrule
DiGress & $0.751$ & 0.3523 & 0.0808 & 0.4085 & 0.1561 \\
GDSS-Prodigy & $0.834$ & 0.5492 & 0.0679 & \textbf{0.0063} & 0.0390 \\
MuDiff & $0.534$ & 0.4948 & 0.9830 & 0.2678 & 0.1189 \\
\midrule
\multicolumn{6}{l}{\textbf{Benchmark 3: Strongly Connected Mesh}} \\
\addlinespace[2pt]
Method &  Reward & \multicolumn{2}{r}{$d_{\mathrm{square}}$} & \multicolumn{2}{r}{$d_{\lambda_2}$} \\
\midrule
\multirow{2}{*}{\textbf{Ours}} & \textbf{0.647} & \multicolumn{2}{r}{0.060} & \multicolumn{2}{r}{\textbf{0.648}} \\
 & $\pm 0.010$ & \multicolumn{2}{r}{$\pm 0.034$} & \multicolumn{2}{r}{$\pm 0.027$} \\
\midrule
HOG & $0.360$ & \multicolumn{2}{r}{0.2962} & \multicolumn{2}{r}{0.9868} \\
Delaunay & $0.578$ & \multicolumn{2}{r}{0.0735} & \multicolumn{2}{r}{0.7699} \\
Plantri & --- & \multicolumn{2}{r}{---} & \multicolumn{2}{r}{---} \\
Boltzmann & $0.502$ & \multicolumn{2}{r}{0.2293} & \multicolumn{2}{r}{0.7647} \\
Erd\H{o}s--R\'enyi & --- & \multicolumn{2}{r}{---} & \multicolumn{2}{r}{---} \\
ERGM & $0.487$ & \multicolumn{2}{r}{0.3459} & \multicolumn{2}{r}{0.6807} \\
Lattice & $0.587$ & \multicolumn{2}{r}{\textbf{0.0000}} & \multicolumn{2}{r}{0.8237} \\
\midrule
DiGress & $0.699$ & \multicolumn{2}{r}{0.4440} & \multicolumn{2}{r}{0.1628} \\
GDSS-Prodigy & $0.386$ & \multicolumn{2}{r}{0.2945} & \multicolumn{2}{r}{0.9355} \\
MuDiff & $0.410$ & \multicolumn{2}{r}{0.6708} & \multicolumn{2}{r}{0.5074} \\
\midrule
\multicolumn{6}{l}{\textbf{Benchmark 4: Graph with 40 Triangles}} \\
\addlinespace[2pt]
Method &  Reward & \multicolumn{2}{r}{$d_{\mathrm{iso}}$} & \multicolumn{2}{r}{$d_{T}$} \\
\midrule
\multirow{2}{*}{\textbf{Ours}} & \textbf{0.979} & \multicolumn{2}{r}{0.042} & \multicolumn{2}{r}{\textbf{0.0011}} \\
 & $\pm 0.022$ & \multicolumn{2}{r}{$\pm 0.043$} & \multicolumn{2}{r}{$\pm 0.0011$} \\
\midrule
HOG & $0.856$ & \multicolumn{2}{r}{0.0571} & \multicolumn{2}{r}{0.2308} \\
Delaunay & $0.759$ & \multicolumn{2}{r}{\textbf{0.0262}} & \multicolumn{2}{r}{0.4558} \\
Plantri & $0.494$ & \multicolumn{2}{r}{0.5244} & \multicolumn{2}{r}{0.4872} \\
Boltzmann & $0.875$ & \multicolumn{2}{r}{0.2247} & \multicolumn{2}{r}{0.0256} \\
Erd\H{o}s--R\'enyi & $0.415$ & \multicolumn{2}{r}{0.1703} & \multicolumn{2}{r}{1.0000} \\
ERGM & $0.333$ & \multicolumn{2}{r}{0.3333} & \multicolumn{2}{r}{1.0000} \\
Lattice & $0.482$ & \multicolumn{2}{r}{0.0352} & \multicolumn{2}{r}{1.0000} \\
\midrule
DiGress & $0.569$ & \multicolumn{2}{r}{0.4065} & \multicolumn{2}{r}{0.4545} \\
GDSS-Prodigy & $0.552$ & \multicolumn{2}{r}{0.3578} & \multicolumn{2}{r}{0.5385} \\
MuDiff & $0.433$ & \multicolumn{2}{r}{0.7050} & \multicolumn{2}{r}{0.4286} \\
\bottomrule
\end{tabular}

\end{table}

%%%%%%%%%%%%%%%%%%%%%%%%%%%%%%%%%%%%%%%%%%%%%%%%%%%%%%%%%%%%
%\input{checklist.tex}

\end{document}